%% file: acl_latex.tex
\documentclass[11pt]{article}

\usepackage[final]{acl}

\usepackage{times}
\usepackage{latexsym}

\usepackage[T1]{fontenc}

\usepackage[utf8]{inputenc}

\usepackage{microtype}

\usepackage{inconsolata}

\usepackage{graphicx}

\usepackage{amsmath}
\usepackage{amssymb}
\usepackage{booktabs}
\usepackage{multirow}
\usepackage{multicol}
\usepackage{enumitem}

\usepackage[table]{xcolor}
\usepackage{tcolorbox}
\tcbuselibrary{skins}

\usepackage{xcolor}
\usepackage{xspace}
\usepackage{CJKutf8}

\newcommand{\codefence}{\texttt{\char96\char96\char96}}

\title{GRRM: Group Relative Reward Modeling for Machine Translation}

\author{
 \textbf{Sen Yang\textsuperscript{1}},
 \textbf{Shanbo Cheng\textsuperscript{1}},
 \textbf{Lu Xu\textsuperscript{2}},
 \textbf{Jianbing Zhang\textsuperscript{1}},
 \textbf{Shujian Huang\textsuperscript{1*}}
\\
 \textsuperscript{1}National Key Laboratory for Novel Software Technology, Nanjing University,
 \\
 \textsuperscript{2}Singapore University of Technology and Design,
\\
\small{
  \texttt{yangsen@smail.nju.edu.cn},\quad\texttt{chengsb@nlp.nju.edu.cn},\quad\texttt{\{huangsj,zjb\}@nju.edu.cn}
}\\
\small{
  \texttt{xu\_lu@mymail.sutd.edu.sg}
}
}

\begin{document}
\maketitle

\renewcommand{\thefootnote}{\fnsymbol{footnote}}
\footnotetext[1]{Corresponding author}
\renewcommand{\thefootnote}{\arabic{footnote}}
\begin{abstract}
While Group Relative Policy Optimization (GRPO) offers a powerful framework for LLM post-training, its effectiveness in open-ended domains like Machine Translation hinges on accurate intra-group ranking. We identify that standard Pointwise Quality Metrics (PQM) fall short in this context: candidates are evaluated in isolation, so the comparative context is missing for distinguishing fine-grained linguistic nuances.
To address this, we introduce the Group Quality Metric (GQM) paradigm and its instantiation, the Group Relative Reward Model (GRRM). Unlike traditional independent scorers, GRRM jointly processes the entire candidate group, leveraging comparative analysis to rigorously resolve relative quality and adaptive granularity.
Empirical evaluations confirm that GRRM achieves competitive ranking accuracy among all baselines; integrating GRRM into the GRPO training not only improves general translation quality but also unlocks reasoning capabilities comparable to state-of-the-art reasoning models.
We release codes, datasets, and model checkpoints at \url{https://github.com/NJUNLP/GRRM}.
\end{abstract}

\input{chapters/chap1}
\input{chapters/chap2}
\input{chapters/chap3}
\input{chapters/chap4}

\input{chapters/chap5}
\input{chapters/chap6}

\section*{Limitations}

We acknowledge several limitations of the proposed Group Quality Metric (GQM) and GRRM.
First, unlike PQM which scales linearly, GQM is constrained by the maximum group size covered during training. Extrapolating to group sizes significantly beyond the training distribution may degrade ranking accuracy. However, this limitation does not hinder effective GRPO training. Recent work demonstrates that using small group sizes combined with larger batch sizes yields performance comparable to larger group configurations~\cite{wu2025takes}, rendering GQM's capacity sufficient for optimization.

Second, GQM relies on relative ranking, which introduces challenges in reward assignment when the overall group quality is low. Since GRRM treats intra-group ranking as the ground truth for advantage estimation, it may assign high relative scores to the "best" candidate even if all translations in the group are suboptimal. Consequently, GQM might fail to penalize the group globally as effectively as an absolute metric would. Integrating absolute quality constraints alongside relative ranking remains a promising direction for future work.

Efficiency is another limitation of GRRM. It is more efficient than pointwise generative reward models for group-wise reward assignment, as it evaluates all candidates in a single generation rather than decoding each evaluation separately. Nevertheless, GRRM still incurs autoregressive decoding costs, while discriminative reward models score candidates with non-autoregressive forward passes. Thus, GRRM improves efficiency within the generative paradigm but remains less efficient than discriminative alternatives.

\section*{Acknowledgments}

We would like to thank the anonymous reviewers for their insightful comments. Shujian Huang is the corresponding author. This work is supported by National Science Foundation of China (No. 62376116), the Fundamental Research Funds for the Central Universities (No. 2024300507), Fundamental and Interdisciplinary Disciplines Breakthrough Plan of the Ministry of Education of China (No. JYB2025XDXM118).

\bibliography{custom,anthology}

\appendix

\input{chapters/appendix}

\end{document}

%% file: chapters/chap1.tex
\section{Introduction}

Reinforcement learning with verifiable rewards (RLVR) has significantly advanced the reasoning processes of Large Language Models (LLMs), yielding remarkable performance in sophisticated domains such as mathematics and programming~\cite{lambert2024tulu,guo2025deepseek}.
Performing reinforcement learning on open-ended tasks, such as Machine Translation (MT), requires reward models to evaluate the responses generated by the policy model.
Existing works predominantly employ Discriminative Reward Models (DRMs) based on the Bradley-Terry model~\cite{cheng2025seed,yang-etal-2025-enanchored}.
While DRMs have proven effective in improving translation quality in general domains, they have been criticized for lacking reasoning capabilities, which limits their effectiveness in challenging scenarios.

\begin{figure}[t]
    \centering
    \includegraphics[width=0.48\textwidth]{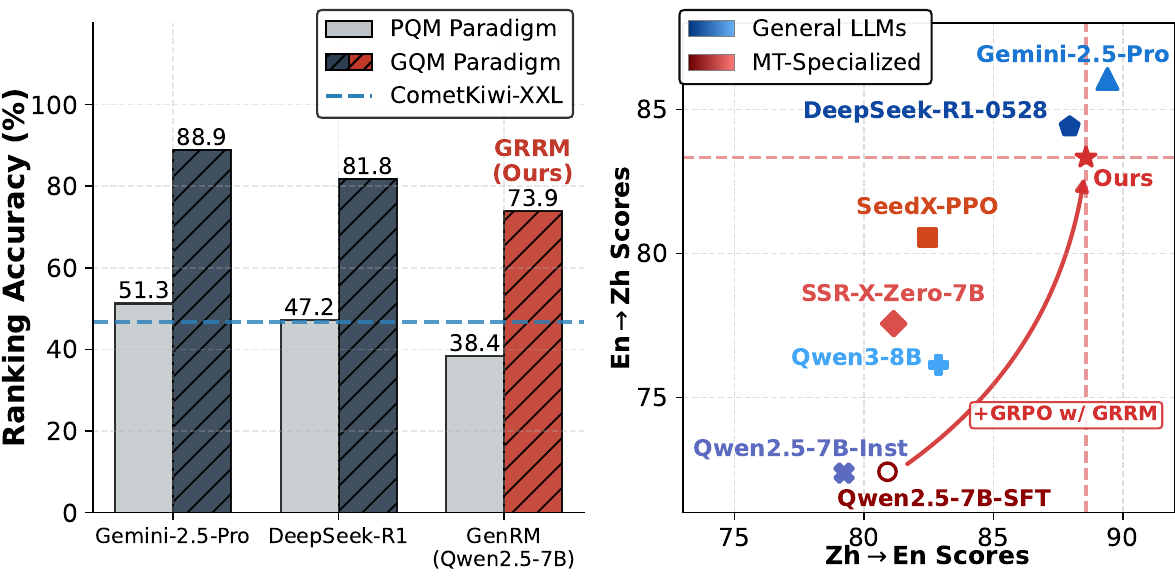}
    \vspace{-1em}
    \caption{Performance on Seed-X-Challenge. \textbf{Left}: Ranking accuracy across paradigms. \textbf{Right}: Translation performance across General and MT-specialized LLMs.}
    \label{fig:teaser}
\end{figure}

Recently, the success of LLM-based generative evaluation (LLM-as-a-Judge) in MT~\cite{kocmi-federmann-2023-gemba,kocmi-federmann-2023-large} suggests the potential of Generative Reward Models~\citep[GenRMs]{zhanggenerative,mahan2024generative}. Ideally, the reasoning potential of GenRMs offers more accurate reward estimation, particularly for difficult samples.
However, we identify a critical limitation when applying GenRMs within the widely used Group Relative Policy Optimization~\citep[GRPO]{shao2024deepseekmath} framework.
Since GRPO estimates advantages based on the relative quality of responses within a group, the reward model must achieve accurate fine-grained intra-group ranking.
Our investigation reveals that standard generative approaches, which evaluate responses independently (\textbf{Pointwise Quality Metric, PQM}), suffer from two fundamental flaws: the inability to distinguish subtle nuances and the vulnerability to misleading translations in challenging samples.
These flaws lead to \textbf{score saturation}, where the model assigns identical maximal scores to candidates despite significant quality disparities. This lack of variance causes vanishing advantages in GRPO, depriving the policy of the gradient signals needed for optimization.

To address the limitations of the PQM paradigm, we propose the \textbf{Group Quality Metric (GQM)}. Specifically, GQM enables the model to process all candidate translations within a group collectively, aiming to jointly estimate their quality ranking and assign relative scores. This cross-sample context provides consistent evaluation criteria, allowing the model to focus on distinguishing features between candidates. Furthermore, this paradigm allows the model to adaptively adjust its evaluation granularity according to the quality variance within the group. 

However, current LLMs are weak when working in GQM style. We present the training pipeline of an instantiation of GQM, i.e. \textbf{Group Relative Reward Model (GRRM)}, a generative reward model structurally aligned with the group-relative nature of GRPO.
Our GRRM is initialized from Qwen2.5-7B~\cite{yang2024qwen2}, cold-started via Supervised Fine-Tuning (SFT) on Chinese-English data, and subsequently optimized for ranking accuracy via RLVR.
We demonstrate that both our trained GRRM and general LLMs evaluated under the GQM formulation consistently outperform their PQM counterparts, verifying the universality of the proposed metric. Notably, on challenging datasets, this paradigm delivers significant absolute gains of 30\% to 40\% in translation quality ranking accuracy compared to PQM (Fig.~\ref{fig:teaser} Left).
Furthermore, our GRRM exhibits strong cross-lingual generalization; despite being trained on a single language pair, it effectively supports multilingual translation optimization without additional adaptation.

We further validate our method in machine translation GRPO training, by using GRRM feedback to optimize the same Qwen2.5-7B backbone.
For general domains, our model achieves an average 7.5-point improvement in BLEURT and a 16 point increase in LLM-judge scores over the SFT baseline across seven English-to-X (En2X) language pairs.
Moreover, in challenging Chinese-English scenarios, our model performs comparably to DeepSeek-R1-0528~\cite{guo2025deepseek} (Fig.~\ref{fig:teaser} Right).
Our analysis shows that GRRM improves the reasoning capabilities of translation policies, which is crucial for solving complex translation challenges. In contrast, while DRMs can improve general translation quality, they are prone to reward hacking and lack the reasoning faculties required to guide models in challenging contexts.



As an additional application, we show that GRRM-based inference-time reranking improves translation quality, including for models already optimized via GRPO; details are provided in Appendix~\ref{app:rm_inference_reranking}.

The remainder of this paper is organized as follows. Section~\ref{sec:methodology} introduces the GQM paradigm and the GRRM training pipeline. Section~\ref{sec:grrm_eval} first evaluates GQM and GRRM as standalone group-wise ranking models, while Section~\ref{sec:mt_opt} then evaluates GRRM as a reward for downstream multilingual MT optimization with GRPO. Additional experiments and analyses are provided in the appendix.

%% file: chapters/chap2.tex
\section{Methodology}
\label{sec:methodology}

\subsection{Ranking Sensitivity in GRPO}
\label{sec:preliminaries}

GRPO is a reinforcement learning paradigm that has proven highly effective for LLM post-training, particularly in complex reasoning domains.

Formally, for each query $x$, the policy $\pi_\theta$ samples a group of outputs $\mathcal{Y} = \{y_1, \dots, y_G\}$. The policy is optimized to maximize the expected advantage of the generations. Specifically, the advantage $A_i$ for the $i$-th candidate is derived by standardizing its reward $r_i$ against the group distribution:
\begin{equation}
    \label{eq:advantage}
    A_i = \frac{r_i - \operatorname{mean}(\{r_1, \dots, r_G\})}{\operatorname{std}(\{r_1, \dots, r_G\})}
\end{equation}
where $r_i$ is the scalar reward score assigned to $y_i$.

Eq.~\ref{eq:advantage} underscores a critical property of GRPO: optimization is driven by the relative quality of candidates within a group, rather than their absolute scores.
A candidate $y_i$ is reinforced ($A_i > 0$) solely if it outperforms the group average. Consequently, GRPO depends on the reward model's ability to accurately rank candidates within $\mathcal{Y}$; any ranking inversion yields adversarial gradients that mislead policy optimization. 
This sensitivity motivates our investigation into the ranking limitations of current reward modeling paradigms.

\subsection{Limitations of Pointwise Quality Metric}
\label{sec:pqm}

\begin{figure}[t]
    \centering
    \includegraphics[width=0.48\textwidth]{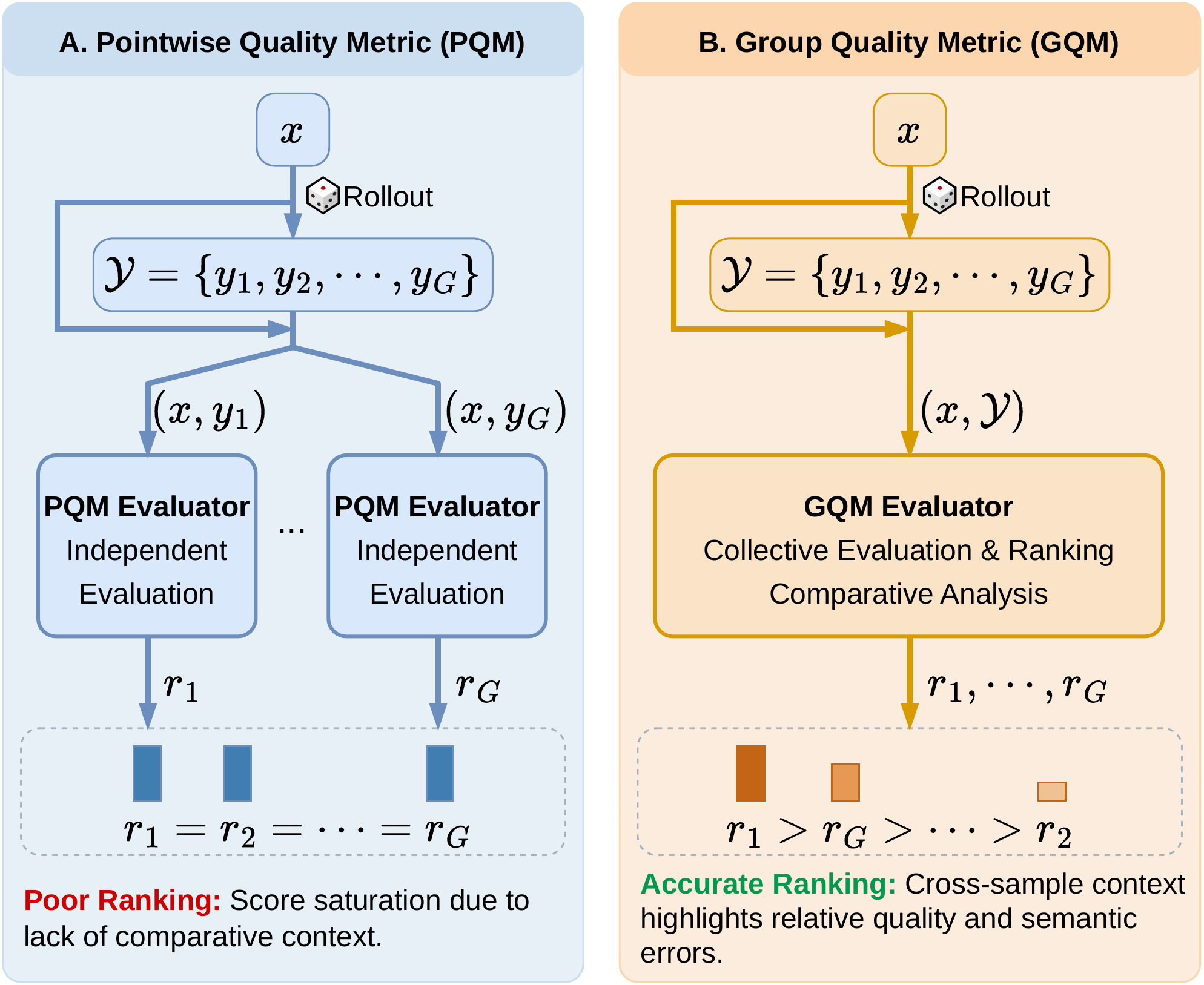}
    \vspace{-1em}
    \caption{Comparison of PQM (A) and GQM (B).}

    \label{fig:paradigm_comparison}
\end{figure}




Current generative reward models typically function as a \textit{Pointwise Quality Metric} (PQM), as illustrated in Figure~\ref{fig:paradigm_comparison} (A).
Formally, PQM is defined as a mapping $S_{\text{PQM}}: \mathcal{X} \times \mathcal{Y} \to \mathbb{R}$ that assigns a scalar score $r_i$ to a candidate $y_i$ given an input $x$. Crucially, this evaluation is performed independently for each candidate. While this pointwise scoring is standard for general quality estimation, our analysis demonstrates that such independence fundamentally limits the model's ability to capture the fine-grained relative distinctions required for effective GRPO updates.

\paragraph{Inability to Distinguish Subtle Differences.}
LLMs frequently generate candidates that differ only in fine-grained nuances. Without comparative context, PQM suffers from discriminative failure. As seen in Appendix Fig.~\ref{fig:pqm_vs_gqm_cases} (Case 1), the model fails to perceive the stylistic superiority of one translation over another, treating them as equivalent.

\paragraph{Vulnerability to Misleading Candidates.}
More critically, independent evaluation leaves the reward model vulnerable to misleading candidates. 
Lacking a high-quality peer for contrast, even powerful models may overlook semantic discrepancies or localization errors, defaulting to surface fluency as proxies for quality. 
As shown in Appendix Fig.~\ref{fig:pqm_vs_gqm_cases} (Case 2), PQM fails to penalize severe localization errors if they appear plausible in isolation.

\paragraph{Score Saturation.}
Both discriminative failure and susceptibility to misleading candidates culminate in a phenomenon we term \textbf{score saturation}. 
PQM frequently assigns identical maximal scores to candidates despite significant quality disparities, causing vanishing advantages in Eq.~\ref{eq:advantage} and stalling policy optimization. 
A detailed empirical analysis is provided in Appendix~\ref{app:saturation_analysis}.

\subsection{Group Quality Metric}
\label{sec:gqm}

To overcome the structural limitations of PQM, we introduce the \textit{Group Quality Metric} (GQM), depicted in Figure~\ref{fig:paradigm_comparison} (B). Unlike the point-wise approach, GQM evaluates the policy-generated candidate group collectively. Formally, it maps a source $x$ and a candidate set $\mathcal{Y}$ to a score vector: $S_{\text{GQM}}: \mathcal{X} \times \mathcal{Y}^G \to \mathbb{R}^G$. This holistic evaluation shifts the paradigm from absolute estimation to \textit{relative ranking}, offering three key advantages:

By evaluating candidates jointly, GQM constructs a comparative context that circumvents the discriminative bottlenecks of PQM. This context mitigates score saturation, enabling the model to resolve fine-grained nuances that remain imperceptible under independent evaluation (Appendix Fig.~\ref{fig:pqm_vs_gqm_cases}, Case 1). Simultaneously, GQM introduces contrastive anchoring: plausible but erroneous translations (e.g., hallucinations) become salient when directly compared against high-quality peers (Appendix Fig.~\ref{fig:pqm_vs_gqm_cases}, Case 2). This mechanism ensures the reward signal captures both subtle stylistic preferences and critical semantic errors.

Crucially, GQM scores are inherently localized and do not map to a global absolute scale. This allows the model to adaptively adjust evaluation granularity: shifting focus to minor stylistic differences when group variance is low, while prioritizing major semantic discrepancies when significant quality divergences exist.

\subsection{Group Relative Reward Model}
\label{sec:grrm}

While the GQM paradigm demonstrates efficacy when powered by frontier LLMs, directly integrating such massive models into the GRPO training loop is computationally intractable due to high latency and inference costs.
We therefore introduce the Group Relative Reward Model (GRRM), an efficient instantiation of the GQM paradigm designed for high-throughput training loops.

GRRM is designed to process the source input $x$ and the full candidate set $\mathcal{Y}$ jointly. For every tuple $(x, \mathcal{Y})$, the model is instructed to produce a structured output comprising: (1) a detailed comparative analysis of the candidates, (2) a predicted ranking, and (3) a set of scalar scores $\{r_1, \dots, r_G\}$ that strictly adhere to the predicted ranking. 
This design enforces a Chain-of-Thought (CoT) reasoning process, ensuring numerical scores are grounded in explicit comparative analysis.
Fig.~\ref{fig:GRRM_example} in Appendix shows a concrete example of GRRM prompting and output.

We initialize GRRM via SFT on a curated dataset to establish instruction adherence and comparative logic. Subsequently, to sharpen the model's discriminative ability to distinguish subtle nuances, we employ RLVR to directly optimize its ranking accuracy.
In this stage, we treat the GRRM itself as the policy to be optimized. To drive this optimization, we define a verifiable reward function, Ranking Accuracy, based on ground-truth orderings. Let $\mathbf{q} = \{q_1, \dots, q_G\}$ be the ground-truth quality scores (or ranks) for the candidate group, and $\mathbf{r} = \{{r}_1, \dots, {r}_G\}$ be the corresponding scores predicted by GRRM. We define the correctness of the pairwise relationship between any two candidates $y_i$ and $y_j$ as:
\begin{equation}
    \mathbb{I}_{ij} = \mathbb{I}\left[\operatorname{sgn}(\hat{r}_i - \hat{r}_j) = \operatorname{sgn}(q_i - q_j)\right]
\end{equation}

where $\operatorname{sgn}(\cdot)$ is the sign function, and $\mathbb{I}(\cdot)$ is the indicator function. This formulation enforces strict alignment, requiring the predicted relationship---including ties---to match the ground truth.
The final ranking accuracy reward $R_{\text{acc}}$ is then calculated as the fraction of correctly ordered pairs over all distinct combinations in the group:
\begin{equation}
    \label{eq:ranking_acc}
    R_{\text{acc}} = \frac{1}{\binom{G}{2}} \sum_{1 \le i < j \le G} \mathbb{I}_{ij}
\end{equation}
This metric functions as a normalized variant of the Kendall's $\tau$ coefficient~\cite{kendall1938new}, adapted to the range $[0, 1]$ with explicit handling of ties.

Additionally, to complement the sparse ranking signal and ensure the predicted scalar scores are well-distributed, we further use an auxiliary score consistency reward, detailed in Appendix~\ref{app:rm_aux_loss}.

%% file: chapters/chap3.tex
\section{GRRM Evaluation}
\label{sec:grrm_eval}

In this section, we provide a comprehensive evaluation of the proposed GRRM within the GQM framework, benchmarking the ranking accuracy (Eq.~\ref{eq:ranking_acc}) against multiple baselines. Our experiments span three distinct settings: in-domain performance, multilingual generalization, and complex reasoning scenarios designed to stress-test reward modeling capabilities.

\subsection{Experimental Setup}
\label{sec:exp_setup}

\paragraph{Datasets and Benchmarks.}
We construct our training data using the Chinese-English subset of TowerBlocks~\cite{alves2024tower}, comprising approximately 18.8k samples. We sample 2-4 translation candidates per source and annotate them with \texttt{Gemini-2.5-Pro}~\cite{comanici2025gemini} under the GQM paradigm. A held-out set of 512 samples is reserved as the model-annotated test set.
For RLVR, we utilize the same underlying data as SFT.

We utilize Newstest2020 (Zh$\to$En) with pSQM (professional Scalar Quality Metric) labels~\cite{freitag2021experts} from the WMT20 metrics task. To evaluate cross-lingual generalization, we employ GeneralMT2022 with MQM (Multidimensional Quality Metric) annotations~\cite{freitag2021experts} from WMT22. Beyond the in-domain Zh$\leftrightarrow$En pairs, this dataset includes English-to-German (En$\to$De) and English-to-Russian (En$\to$Ru) subsets. For these datasets, we retain up to 4 system outputs per source and derive ground-truth rankings based on the human-annotated scores.

To assess reasoning capabilities in complex linguistic scenarios, we construct a challenge set based on the Seed-X-Challenge test set~\cite{cheng2025seed}, which features idioms, slang, and domain-specific terminology in Zh$\leftrightarrow$En pairs.
We design a binary ranking task to test the model's reasoning capability: for each source, we sample 4 translation candidates and select the best one according to BLEURT~\cite{sellam-etal-2020-bleurt}. We then pair this chosen candidate with the expert human reference.
Assuming the expert reference is strictly superior to the weak baseline's output, this setup requires the reward model to correctly identify the human translation as the winner.

\paragraph{Baselines.}
We include three categories of reward models for comparison. Unless otherwise noted, all trained models are initialized from \texttt{Qwen2.5-7B}.
\begin{itemize}[noitemsep]
    \item \textbf{LLM-as-a-Judge:} We test \texttt{Gemini-2.5-Pro} and \texttt{DeepSeek-R1-0528}. We evaluate them using both the standard PQM prompting and our proposed GQM prompting, explicitly instructing the models to use CoT reasoning.
    \item \textbf{DRMs:} We include CometKiwi-XXL~\cite{rei-etal-2023-scaling}, a leading reference-free Quality Estimation metric. Additionally, we train a BT-RM using the standard Bradley-Terry loss on pairwise preferences derived from our Gemini-annotated training data.
    \item \textbf{GenRMs:} In addition to our GRRM, we train a PQM-GenRM baseline using Gemini PQM annotations, following an RLVR optimization process similar to that of GRRM.
\end{itemize}
See Appendix~\ref{app:imp_rm_training} for details on data construction and training hyperparameters.

\begin{table*}[t]
\centering
\small
\setlength{\tabcolsep}{3.2pt}
\begin{tabular}{l c c c c c c c c c}
\toprule
\multirow{2}{*}{\textbf{Model}} & \multirow{2}{*}{\textbf{Paradigm}} & \textbf{held-out} & \textbf{NT20} & \multicolumn{4}{c}{\textbf{GenMT22 (MQM)}} & \textbf{Seed-X-Challenge} & \multirow{2}{*}{\textbf{Avg.}} \\
\cmidrule(lr){3-3} \cmidrule(lr){4-4} \cmidrule(lr){5-8} \cmidrule(lr){9-9}
 & & Zh$\leftrightarrow$En & Zh$\to$En & En$\to$Zh & Zh$\to$En & En$\to$De & En$\to$Ru & Zh$\leftrightarrow$En & \\
\midrule
Random & - & 43.47 & 39.42 & 43.82 & 38.74 & 35.91 & 37.64 & 44.70 & 40.04 \\
\midrule
\multicolumn{10}{l}{\textit{LLM-as-a-Judge (w/ Reasoning)}} \\
Gemini-2.5-Pro & PQM & 70.28 & 53.56 & 46.41 & 61.99 & 58.92 & 64.88 & 51.26 & 56.17 \\
Gemini-2.5-Pro & GQM & - & \textbf{62.60} & \textbf{64.58} & \textbf{67.75} & \textbf{65.61} & \textbf{71.23} & \textbf{88.89} & \textbf{70.11} \\
DeepSeek-R1-0528 & PQM & 66.11 & 48.42 & 43.67 & 58.09 & 53.64 & 59.13 & 47.22 & 51.69 \\
DeepSeek-R1-0528 & GQM & \textbf{80.92} & 61.98 & 64.38 & 65.79 & 63.26 & 69.15 & 81.82 & 67.73 \\
\midrule
\multicolumn{10}{l}{\textit{Discriminative RMs (w/o Reasoning)}} \\
CometKiwi-XXL & PQM & 72.01 & 57.82 & \textbf{66.49} & 61.60 & \textbf{61.20} & \textbf{67.12} & 46.72 & 60.16 \\
BT-RM & PQM & \textbf{82.62} & \textbf{58.16} & \textbf{66.49} & \textbf{64.92} & 58.14 & 66.10 & 58.84 & \textbf{62.11} \\
\midrule
\multicolumn{10}{l}{\textit{Generative RMs (w/ Reasoning)}} \\
PQM-GenRM (SFT) & PQM & 61.31 & 47.18 & 37.21 & 57.68 & 48.37 & 51.84 & 38.13 & 46.74 \\
PQM-GenRM (RLVR) & PQM & 64.25 & 49.21 & 39.42 & 60.37 & 49.16 & 54.82 & 38.38 & 48.56 \\
\textbf{GRRM} (SFT) & GQM & 79.75 & 56.77 & 59.45 & 64.29 & 58.65 & 64.80 & 69.95 & 62.32 \\
\textbf{GRRM} (RLVR) & GQM & \textbf{83.19} & \textbf{57.57} & \textbf{62.12} & \textbf{65.55} & \textbf{60.75} & \textbf{66.39} & \textbf{73.93} & \textbf{64.38} \\
\bottomrule
\end{tabular}
\caption{Ranking accuracy (\%) on the \textbf{held-out} test set (Gemini-annotated) versus human-annotated benchmarks and the challenge set. \textbf{Avg.} denotes the mean accuracy across all external benchmarks (excluding held-out). 
The best performance within each model category is \textbf{bolded}.}

\label{tab:ranking_acc_results}
\end{table*}

\subsection{Performance of GRRM}
\label{sec:ranking_acc_results}

Table~\ref{tab:ranking_acc_results} presents the ranking accuracy of various reward models across held-out, general, and challenging datasets. We summarize our key findings as follows:

\paragraph{Improvements from GQM over PQM.}
GQM improves ranking accuracy over the corresponding PQM setup across evaluator backbones, with the largest gains in the challenge scenario.
For LLM-as-a-Judge, employing GQM yields substantial improvements; for instance, \texttt{Gemini-2.5-Pro} achieves an average accuracy of 70.11\% with GQM, compared to just 56.17\% with PQM.
Notably, this performance gap is maximized in the challenge scenario. While PQM performance for even state-of-the-art models degrades to near-random levels (e.g., 47.22\% for DeepSeek-R1), GQM maintains robust accuracy.
This trend holds for trained reward models as well, where GRRM (based on GQM) significantly outperforms PQM-GenRM.
These results validate that independent evaluation (PQM) fails to capture fine-grained relative differences. By processing candidates collectively, GQM provides the model with a comparative context, enabling more robust and consistent ranking.

\paragraph{Effectiveness of GRRM and Reasoning.}
Our proposed GRRM achieves the highest average accuracy among all trained models.
While we observe that RLVR optimization brings consistent performance gains over the SFT cold-start, the magnitude of improvement is less significant than the paradigm shift from PQM to GQM.
Crucially, the advantage of GRRM is most pronounced in the Seed-X Challenge set, achieving a ranking accuracy of 73.93\%.
This indicates that the reasoning capability is essential for verifying complex linguistic phenomena that statistical correlations alone cannot capture.
Additionally, despite being trained solely on Zh-En data, GRRM exhibits strong cross-lingual generalization to unseen En$\to$De and En$\to$Ru pairs, performing on par with the CometKiwi-XXL baseline.

\paragraph{Limitations of Discriminative RMs.}
DRMs perform competitively on general domain benchmarks. This can be attributed to the Bradley-Terry loss, which implicitly and effectively models the ranking objective on preference pairs. However, their performance collapses on the challenge set due to the lack of generative reasoning capabilities, limiting their utility in guiding policies through complex optimization landscapes.
For additional comparisons against discriminative baselines that incorporate limited contrastive context, see Appendix~\ref{app:contrastive_discriminative_baselines}.

%% file: chapters/chap4.tex
\section{Applying GRRM to MT Optimization}
\label{sec:mt_opt}

Building upon the validation of GRRM's ranking capabilities, we integrate it into the training loop to optimize the translation policy. This section evaluates whether GRRM-guided optimization improves MT performance and translation-relevant reasoning capabilities.

\subsection{Experimental Setup}

\paragraph{Training Pipeline.}
We adopt a two-stage strategy initialized with \texttt{Qwen2.5-7B}. First, we perform SFT to set up preliminary translation and reasoning skills using the same Chinese-English data described in Section~\ref{sec:grrm_eval} (with \texttt{Gemini-2.5-Pro} annotated CoT). Second, we optimize the model using GRPO with GRRM feedback based on the multilingual translation data from TowerBlocks, covering 10 languages with approximately 150k samples.

Since GRRM yields reference-free rewards, we implement Cross-Lingual Augmentation (CLA) by pairing source sentences with alternative target languages (e.g., Zh$\to$De) from the dataset, while maintaining the total number of update steps constant.

\paragraph{Datasets.}
Our primary evaluation targets Zh$\leftrightarrow$En translation using WMT23 Zh$\to$En~\cite{kocmi-etal-2023-findings}, WMT24++ En$\to$Zh~\cite{deutsch2025wmt24expandinglanguagecoverage}, and the Seed-X-Challenge. We additionally evaluate En$\to$De, Es, Fr, It, Nl, Pt, and Ru on WMT24++, and De$\to$En, Ja$\to$En, and Ru$\to$En on WMT23. For En$\to$X, we report average performance across the seven WMT24++ languages. Detailed results are provided in Appendix~\ref{app:mt_full_results}.

\paragraph{Evaluation Metrics.}
We employ a combination of standard metrics and LLM-based evaluation to provide a comprehensive assessment:
\begin{itemize}[noitemsep]
    \item \textbf{BLEURT:} We use BLEURT-20 as our primary automatic metric to measure semantic preservation against human references.
    \item \textbf{LLM-as-a-Judge:} Despite our findings in Section~\ref{sec:grrm_eval} regarding the superiority of GQM for ranking, we utilize the PQM paradigm for system-level evaluation. We justify this choice based on two key factors:
    (1) \textbf{Reference Anchoring:} Our evaluation here includes human references. The reference serves as a strong anchor (conceptually similar to a group size of 2 in GQM), significantly stabilizing the judgment.
    (2) \textbf{System-Level Accuracy:} While PQM struggles with fine-grained intra-group ranking, it remains reliable for aggregating scores at the system level.
    
    For Seed-X-Challenge, we further enhance the evaluation prompt by including the expert annotations provided with the dataset. These annotations highlight specific translation difficulties, enabling the judge to perform a more informed assessment alongside the reference.
    
    \item \textbf{Evaluator Models:} To balance overhead, we employ \texttt{DeepSeek-R1-0528} as the judge for all main results and \texttt{gpt-oss-120b} for ablation studies and supplementary results. And we additionally validate the Seed-X-Challenge comparison with newly released \texttt{GPT-5.6-Luna} and \texttt{DeepSeek-v4-Pro} evaluators in Appendix~\ref{app:independent_evaluators}.
\end{itemize}

We also report additional reference-based results using MetricX-24 in Appendix~\ref{app:metricx24_results}.

\paragraph{Baselines.}
We incorporate the following baselines:
\begin{itemize}[noitemsep]
    \item \textbf{General LLMs:} We test \texttt{Gemini-2.5-Pro} and \texttt{DeepSeek-R1-0528}, along with \texttt{Qwen3-8B}~\cite{yang2025qwen3} and \texttt{Qwen2.5-7B-Instruct}~\cite{yang2024qwen2} to benchmark improvements against the base model family.
    
    \item \textbf{Translation-Specialized Models:} We compare against three state-of-the-art open models:
    (1) \texttt{TowerInstruct-13B}~\citep{alves2024tower}, a LLaMA2~\cite{touvron2023llama} derivative specifically adapted for translation tasks through continued pre-training and instruction tuning;
    (2) \texttt{SeedX-PPO}~\citep{cheng2025seed}, a 7B model pre-trained on high-quality multilingual datasets and subsequently enhanced via RL;
    and (3) \texttt{SSR-X-Zero-7B}~\citep{yang2025ssr}, a \texttt{Qwen2.5-7B} derivative which utilizes a self-rewarding RL framework specifically optimized for Zh$\leftrightarrow$En translation tasks.

\end{itemize}

\paragraph{Reasoning Configuration.}
For General LLMs, we explicitly prompt them to use CoT reasoning to maximize their potential. For specialized models, \texttt{SeedX-PPO} and \texttt{SSR-X-Zero-7B} leverage reasoning, while \texttt{TowerInstruct-13B} supports direct translation only.

\begin{table*}[t]
\centering
\small
\setlength{\tabcolsep}{3.2pt}
\begin{tabular}{lcccccccccc}
\toprule
\multirow{3}{*}{\textbf{Model}} & \multicolumn{6}{c}{\textbf{WMT Benchmarks}} & \multicolumn{4}{c}{\textbf{Seed-X-Challenge}} \\
\cmidrule(lr){2-7} \cmidrule(lr){8-11}
 & \multicolumn{2}{c}{Zh$\to$En} & \multicolumn{2}{c}{En$\to$Zh} & \multicolumn{2}{c}{En$\to$X} & \multicolumn{2}{c}{Zh$\to$En} & \multicolumn{2}{c}{En$\to$Zh} \\
\cmidrule(lr){2-3} \cmidrule(lr){4-5} \cmidrule(lr){6-7} \cmidrule(lr){8-9} \cmidrule(lr){10-11}
 & \scriptsize{BLEURT} & \scriptsize{R1-judge} & \scriptsize{BLEURT} & \scriptsize{R1-judge} & \scriptsize{BLEURT} & \scriptsize{R1-judge} & \scriptsize{BLEURT} & \scriptsize{R1-judge} & \scriptsize{BLEURT} & \scriptsize{R1-judge} \\
\midrule
\multicolumn{11}{l}{\textit{\textbf{General LLMs}}} \\
Gemini-2.5-Pro & \textbf{68.66} & \textbf{92.92} & \textbf{66.00} & \textbf{91.31} & \textbf{68.87} & \textbf{90.35} & \textbf{71.59} & \textbf{89.41} & \textbf{69.19} & \textbf{86.06} \\
DeepSeek-R1-0528 & 67.78 & 92.34 & 64.87 & 89.24 & 67.72 & 88.48 & 70.92 & 87.95 & 68.23 & 84.40 \\
Qwen3-8B & 63.03 & 89.72 & 57.58 & 84.15 & 57.25 & 77.37 & 61.17 & 82.88 & 57.78 & 76.12 \\
Qwen2.5-7B-Instruct & 67.31 & 88.49 & 59.92 & 80.51 & 58.72 & 72.51 & 66.59 & 79.23 & 62.75 & 72.37 \\
\midrule
\multicolumn{11}{l}{\textit{\textbf{Translation-Specialized Models}}} \\
TowerInstruct-13B & 67.56 & 84.83 & 62.92 & 77.63 & 66.61 & 82.68 & 63.32 & 69.54 & 63.46 & 71.17 \\
SeedX-PPO & \textbf{69.02} & 90.47 & \textbf{67.21} & 87.98 & \textbf{68.35} & \textbf{86.04} & 69.37 & 82.47 & \textbf{68.72} & 80.56 \\
SSR-X-Zero-7B & 68.30 & 88.67 & 66.12 & 83.78 & - & - & 68.84 & 81.15 & 67.08 & 77.56 \\
\rowcolor{gray!10} Qwen2.5-7B-SFT & 67.07 & 87.78 & 59.99 & 76.98 & 57.14 & 67.91 & 67.65 & 80.91 & 62.36 & 72.42 \\
\rowcolor{gray!10} \quad + GRPO (ours) & 67.41 & \textbf{92.24} & 64.80 & 87.80 & 64.65 & 83.86 & \textbf{69.55} & 85.90 & 67.05 & 82.55 \\
\rowcolor{gray!10} \quad + GRPO w/ CLA (ours) & 67.39 & 92.09 & 63.91 & \textbf{88.29} & 64.50 & 83.71 & 69.25 & \textbf{88.58} & 67.07 & \textbf{83.33} \\
\bottomrule
\end{tabular}
\caption{\textbf{MT performance on WMT and Seed-X-Challenge benchmarks.} We report BLEURT-20 and LLM-as-a-Judge scores evaluated by \texttt{DeepSeek-R1-0528}. The best performance within each category is highlighted in \textbf{bold}.}
\label{tab:mt_results}
\end{table*}

\subsection{Main Results}

Table~\ref{tab:mt_results} presents the performance of our GRRM-optimized models compared to the SFT baseline and other state-of-the-art systems. The results indicate that integrating GRRM into the GRPO framework significantly enhances translation quality, particularly in scenarios requiring reasoning.

Compared to the \texttt{Qwen2.5-7B-SFT} baseline, our method achieves substantial improvements across all language pairs.
In the general domain (WMT benchmarks), the model achieves an average gain of \textbf{+7.5 BLEURT} points and \textbf{+15.9 LLM-judge} points on En$\to$X tasks.
This successful optimization on multilingual data validates the strong cross-lingual generalization of GRRM, consistent with our metric analysis in Section~\ref{sec:grrm_eval}.
Furthermore, on Zh$\leftrightarrow$En tasks, our approach comprehensively surpasses all \textit{Translation-Specialized Models} in LLM-judge scores, narrowing the gap with much larger proprietary models like \texttt{Gemini-2.5-Pro}.

The advantages of our approach are most pronounced in the challenge scenarios, i.e., Seed-X-Challenge.
Our model equipped with Cross-Lingual Augmentation (CLA) achieves parity with the powerful reasoning model \texttt{DeepSeek-R1-0528}, even though the latter serves as our evaluator.
Specifically, in the Zh$\to$En subset, our model slightly outperforms \texttt{DeepSeek-R1-0528} (88.58 vs. 87.95), while remaining highly competitive in the En$\to$Zh subset (83.33 vs. 84.40).
We provide detailed case studies in Appendix~\ref{app:mt_reasoning_cases} to qualitatively demonstrate how our model successfully navigates these challenging samples with reasoning. A quantitative analysis of the reasoning traces in Appendix~\ref{app:reasoning_trace_analysis} further shows that GRRM optimization improves the accuracy of analyzing expert-identified translation issues.

Our results also reveal a notable divergence between BLEURT and LLM-as-a-Judge scores.
For instance, while \texttt{SeedX-PPO} achieves the highest BLEURT scores in several settings, its performance drops significantly under the scrutiny of the reasoning-based LLM judge. 
We further examine and explain this divergence using two additional independent judges in Appendix~\ref{app:independent_evaluators}.
Their judgments reinforce our observation in Section~\ref{sec:grrm_eval} that traditional discriminative metrics such as BLEURT can miss critical errors in challenging translation scenarios.

\begin{table*}[t]
\centering
\small 
\setlength{\tabcolsep}{3.2pt}
\begin{tabular}{lcccccccccc}
\toprule
\multirow{3}{*}{\textbf{Ablation Configuration}} & \multicolumn{6}{c}{\textbf{WMT Benchmarks}} & \multicolumn{4}{c}{\textbf{Seed-X-Challenge}} \\
\cmidrule(lr){2-7} \cmidrule(lr){8-11}
 & \multicolumn{2}{c}{Zh$\to$En} & \multicolumn{2}{c}{En$\to$Zh} & \multicolumn{2}{c}{En$\to$X} & \multicolumn{2}{c}{Zh$\to$En} & \multicolumn{2}{c}{En$\to$Zh} \\
\cmidrule(lr){2-3} \cmidrule(lr){4-5} \cmidrule(lr){6-7} \cmidrule(lr){8-9} \cmidrule(lr){10-11}
 & \scriptsize{BLEURT} &\scriptsize{oss-judge} & \scriptsize{BLEURT} &\scriptsize{oss-judge} & \scriptsize{BLEURT} &\scriptsize{oss-judge} & \scriptsize{BLEURT} &\scriptsize{oss-judge} & \scriptsize{BLEURT} &\scriptsize{oss-judge} \\
\midrule
\multicolumn{11}{l}{\textit{\textbf{Comparison of Reward Models}}} \\
SFT baseline (w/ Reasoning) & 67.07 & 85.50 & 59.99 & 76.80 & 57.14 & 65.71 & 67.65 & 80.76 & 62.36 & 74.30 \\
\quad + GRPO w/ BLEURT & \textbf{68.71} & 88.21 & \textbf{67.09} & 85.52 & \textbf{66.55} & 82.00 & 69.69 & 83.35 & \textbf{68.66} & 81.85 \\
\quad + GRPO w/ COMETKiwi & 68.69 & 88.78 & 65.28 & 84.50 & 61.44 & 75.40 & 69.04 & 82.64 & 66.82 & 79.39 \\
\quad + GRPO w/ BT-RM & 68.06 & 88.86 & 60.20 & 76.99 & 56.53 & 64.93 & \textbf{69.74} & 85.60 & 58.25 & 66.91 \\
\quad + GRPO w/ PQM-GenRM & 68.63 & 89.01 & 44.28 & 45.54 & 27.53 & 8.15 & 69.85 & 84.25 & 47.19 & 46.42 \\
\rowcolor{gray!10} \quad + GRPO w/ GRRM & 67.41 & \textbf{89.85} & 64.80 & \textbf{87.91} & 64.65 & \textbf{83.86} & 69.55 & \textbf{87.30} & 67.05 & \textbf{84.36} \\
\midrule
\multicolumn{11}{l}{\textit{\textbf{Cross-Lingual Generalization}}} \\
\quad + GRPO w/ GRRM (ZhEn) & 66.53 & 89.11 & 63.07 & 86.88 & 61.70 & 79.18 & 69.26 & 86.16 & 66.20 & 83.86 \\
\midrule
\multicolumn{11}{l}{\textit{\textbf{Role of Reasoning}}} \\
SFT baseline (w/o Reasoning) & 67.61 & 85.28 & 63.50 & 79.42 & 63.24 & 75.33 & 65.52 & 76.83 & 64.04 & 73.34 \\
\quad + GRPO w/ GRRM & 67.38 & 88.42 & 64.15 & 85.19 & 63.40 & 78.88 & 68.57 & 85.38 & 65.92 & 79.35 \\
\bottomrule
\end{tabular}
\caption{\textbf{Ablation study on Reward Models and Reasoning configurations.} We compare different reward signals and training setups. We report BLEURT-20 and LLM-as-a-Judge scores evaluated by \texttt{gpt-oss-120b}.}
\label{tab:ablation}
\end{table*}

\subsection{Ablation Study}
\label{sec:mt_ablation}

We conduct ablation studies focusing on three key aspects: the choice of reward signals, the impact of training data distribution, and the role of reasoning in the policy model (see Table~\ref{tab:ablation}). 
We place extended analyses on the non-reasoning baseline and \texttt{Qwen2.5-7B-Instruct} in Appendix~\ref{app:mt_ext_ablation}.

\paragraph{Comparison of Reward Models.}
Optimizing directly against BLEURT yields the highest scores on the BLEURT metric itself, as expected. However, this gain does not translate effectively to LLM-judge scores, particularly on the Seed-X-Challenge (e.g., 81.85 vs. 84.36 on En$\to$Zh). The same observation holds for CometKiwi~\cite{rei-etal-2022-cometkiwi}\footnote{To reduce training overhead, we use a smaller CometKiwi checkpoint for this ablation rather than the CometKiwi-XXL model used for ranking evaluation in Section~\ref{sec:grrm_eval}.}. This suggests that discriminative metrics fail to guide the model through the complex reasoning processes required for challenging translations.

More critically, we observe severe reward hacking with BT-RM and PQM-GenRM. These models perform catastrophically on En$\to$Zh and En$\to$X tasks
and suffer from off-target generation issues: apt to generate English responses for non-English targets. 
This indicates that standard discriminative or pointwise generative rewards struggle to distinguish instruction-following failures from translation quality issues, allowing the policy to exploit the reward model.
In contrast, GRRM provides robust feedback, effectively penalizing such deviations and preventing reward hacking.

\paragraph{Cross-Lingual Generalization}

We examine whether the cross-lingual generalization observed in our reward model (Section~\ref{sec:grrm_eval}) transfers to Policy Optimization.
We train a variant using only Chinese-English data, i.e., \texttt{GRPO w/ GRRM (ZhEn)}.
While it improves performance on the Zh$\leftrightarrow$En pair, its performance on En$\to$X lags significantly behind the model trained on multilingual data (79.18 vs. 83.86).
This highlights a distinct difference between evaluation and generation: while the Reward Model demonstrates robust cross-lingual generalization, the Translation Policy exhibits significant transfer loss on unseen languages, indicating that explicit multilingual exposure remains crucial for optimal generation quality.

\paragraph{The Necessity of Reasoning.}
Finally, we assess the contribution of the reasoning process (CoT) to translation quality.
It is evident that models without reasoning capabilities consistently underperform their reasoning counterparts.
Even when optimized with GRRM, the non-reasoning model scores lower on both general benchmarks and challenge sets (e.g., 79.35 vs. 84.36 on Seed-X En$\to$Zh).
This validates the necessity of a ``Reasoning $\times$ Reasoning'' paradigm: the full potential of our framework is unlocked only when a reasoning-capable policy is paired with a reasoning-based reward model (GRRM), enabling the system to effectively plan, self-correct, and evaluate in complex scenarios.

%% file: chapters/chap5.tex
\section{Related Work}

\paragraph{Reasoning for Machine Translation}

The integration of reasoning capabilities into Machine Translation has emerged as a promising direction~\cite{chen2025evaluating,liu2025new}.
Previous attempts to integrate reasoning into MT primarily relied on prompting or SFT \citep{feng-etal-2025-tear, wang-etal-2025-drt, cheng2025seed}. However, \citet{zebaze2025llm} observed that fine-tuning on synthetic CoT data alone often fails to outperform standard fine-tuning. 
Fundamentally, these SFT-based methods are not only constrained by the scarcity of high-quality supervision but also struggle to cultivate intrinsic reasoning capabilities.

Inspired by the success of reinforcement learning in mathematical and coding domains, recent works have adopted RL to encourage self-emergent reasoning. \citet{he2025r1} and \citet{feng-etal-2025-mt-r1} adapt the GRPO framework using discriminative metrics like Comet~\cite{rei-etal-2020-comet} and CometKiwi~\cite{rei-etal-2023-scaling} to guide the model. Similarly, SSR-Zero~\cite{yang2025ssr} employs the model itself to assign pointwise quality scores combined with Comet to estimate advantages, demonstrating that self-evaluation can drive policy improvement.
By contrast, ExTrans~\cite{wang2025extrans} leverages a superior large reasoning model to generate exemplar references and evaluates policy outputs by contrasting them against these anchors.
However, these works employ reward mechanisms that are suboptimal for fostering robust reasoning. DRM-based methods lack the generative explanatory power to guide models through complex errors and are prone to reward hacking. The pointwise evaluation mechanism employed by SSR-Zero fails to capture the fine-grained relative rankings within a group.

\paragraph{Comparative and Reasoning-based Reward Modeling}

While machine translation evaluation has traditionally relied on PQM paradigm, employing pairwise comparative paradigms for judgment and reward estimation is not new in general open-ended domains~\cite{li2024generative,zhujudgelm}.
\citet{ye2025learning} leverages self-generated contrastive pairs to train judges via DPO, enhancing robustness against bias compared to pointwise models.
\citet{liu2025pairjudge} introduces a pairwise judge combined with a knockout tournament for Best-of-N sampling.
Furthermore, integrating CoT into evaluation has proven critical; \citet{guo2025reward} demonstrates that executing a deliberate reasoning process before scoring improves accuracy by utilizing test-time compute.

However, these comparative approaches have yet to be effectively adapted to the group-wise context required by the GRPO framework.
Existing pairwise methods are structurally misaligned with GRPO: estimating advantages for a group of responses via pairwise judges necessitates tournament-style evaluations (e.g., ELO ratings), which incur prohibitive computational costs.
In contrast, our GRRM extends the comparative intuition to a list-wise context.
By ranking all candidates in a single generation, our approach avoids the repeated decoding required by pointwise generative reward models and is therefore more efficient for group-wise reward assignment.

%% file: chapters/chap6.tex
\section{Conclusion}

In this work, we identify and address the limitations of pointwise generative judges in GRPO training, specifically their insufficient sensitivity to distinguish intra-group quality differences in machine translation. To overcome this, we propose the Group Quality Metric (GQM), a paradigm that evaluates candidates jointly to capture fine-grained distinctions often missed by pointwise metrics.
Building on this foundation, we introduce the Group Relative Reward Model (GRRM). Our experiments demonstrate that GRRM significantly outperforms pointwise generative baselines, achieving robust ranking accuracy across diverse languages and challenging reasoning tasks. Crucially, integrating GRRM into the GRPO loop strengthens the reasoning capabilities of translation policies. This enables our approach to rival state-of-the-art systems like DeepSeek-R1 in complex scenarios while effectively mitigating reward hacking.

%% file: chapters/appendix.tex
\section{Empirical Analysis of Score Saturation}
\label{app:saturation_analysis}

\begin{figure*}[t]
    \centering
    \includegraphics[width=1\linewidth]{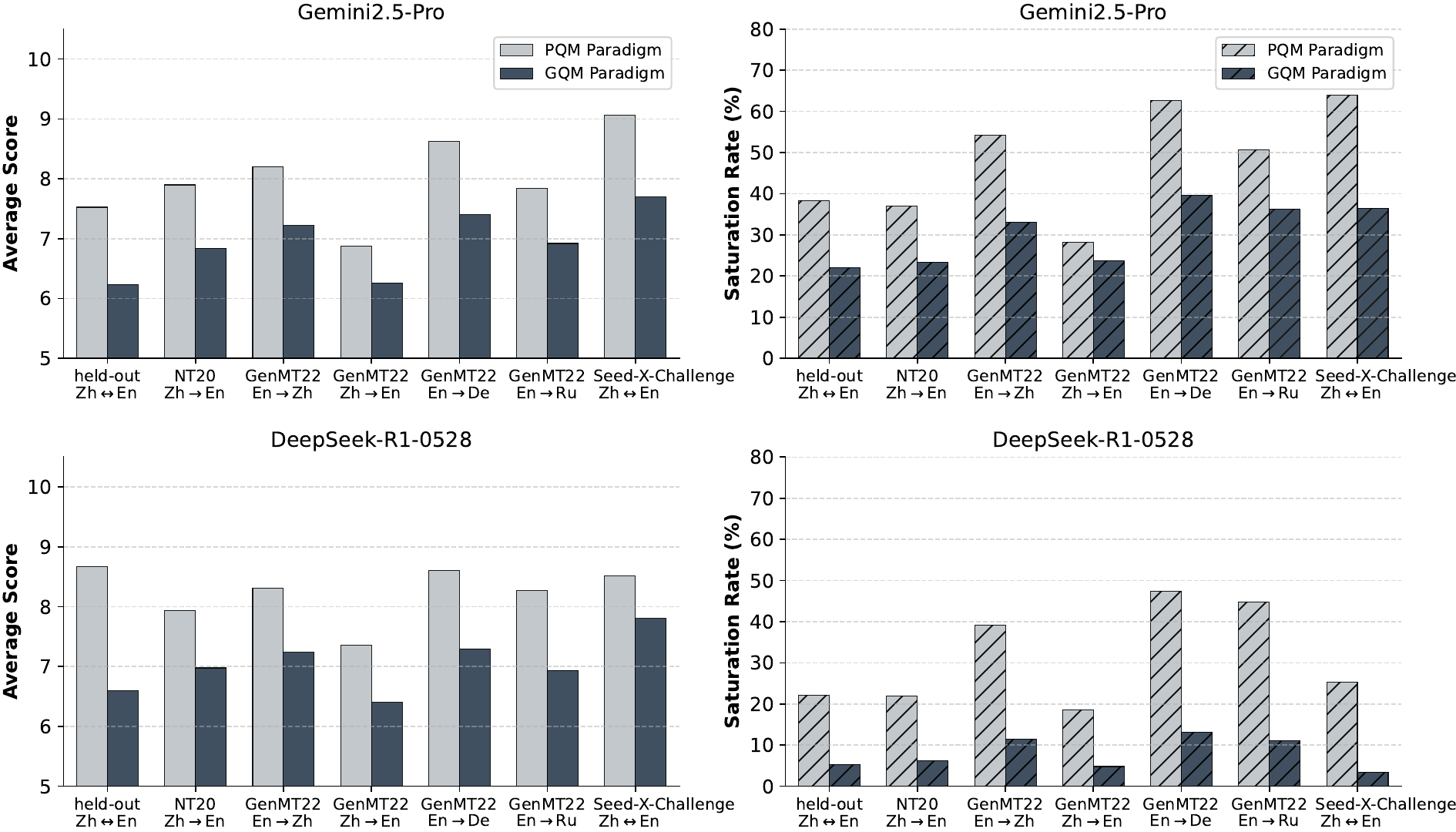}
    \vspace{-1em}
    \caption{\textbf{Analysis of score saturation across seven benchmarks.} We compare the \textbf{Average Score} (left) and \textbf{Saturation Rate} (right) under PQM and GQM paradigms using \texttt{Gemini-2.5-Pro} and \texttt{DeepSeek-R1-0528}. PQM consistently exhibits score inflation and high saturation, whereas GQM effectively mitigates this issue, providing more discriminative evaluation.}
    \label{fig:llm_score_saturation}
\end{figure*}

\begin{figure}[t]
    \raggedright
    \includegraphics[width=0.95\linewidth]{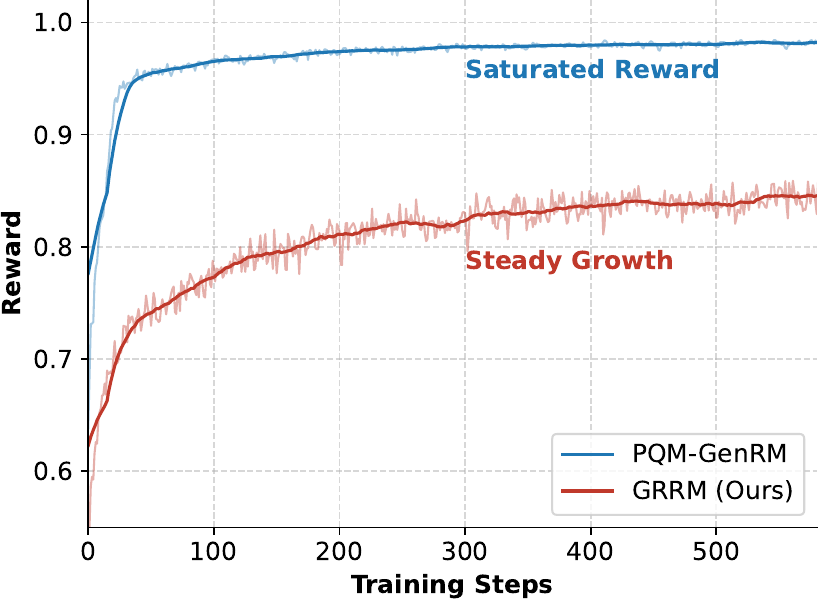}
    \caption{\textbf{Reward trends during GRPO training with 30-step moving averages, using PQM-GenRM vs.\ GRRM as reward providers (scores normalized to $[0,1]$).} The PQM-GenRM rewards saturate early whereas GRRM remains in a non-saturated regime and continues to provide discriminative training signal.}
    \label{fig:rm_reward_curve}
\end{figure}

As discussed in Section~\ref{sec:pqm}, when a model is used under the standard Pointwise Quality Metric (PQM) paradigm, it often produces overly optimistic scores and, in particular, frequently assigns the maximum possible score.
We refer to this phenomenon as \textbf{score saturation}. This reduces reward variance within a candidate group, shrinking the standardized advantages and erasing the learning signal.

In this section, we present an empirical analysis to substantiate our claim that PQM suffers from severe score saturation compared with GQM, which in turn impedes effective policy optimization.

\subsection{Setup}

Our datasets follow Section~\ref{sec:exp_setup}: the held-out Zh$\leftrightarrow$En set, NT20 (Zh$\rightarrow$En), GenMT22 (En$\rightarrow$Zh, Zh$\rightarrow$En, En$\rightarrow$De, En$\rightarrow$Ru), and Seed-X-Challenge (Zh$\leftrightarrow$En).

\paragraph{Saturation in LLM-based evaluation.}

For each dataset, we compute (i) the average score and (ii) the saturation rate, i.e., the fraction of candidates receiving the \emph{maximum allowed score} under the corresponding prompting protocol.
We report results for two advanced LLM judges (i.e., \texttt{Gemini-2.5-Pro} and \texttt{DeepSeek-R1-0528}) under both PQM and GQM prompting.

\paragraph{Saturation during GRPO optimization.}
To connect saturation to optimization dynamics, we also track the reward values observed during GRPO training of the translation policy when using PQM-GenRM versus our GRRM as the reward provider.

\subsection{Results}
\paragraph{PQM yields systematic saturation.}

As shown in Figure~\ref{fig:llm_score_saturation}, across all seven datasets and for both judges, PQM consistently produces higher average scores than GQM, accompanied by markedly higher saturation rates.
For example, with \texttt{Gemini-2.5-Pro}, PQM saturation ranges from 28.24\% to 64.02\%, while GQM reduces it to 22.04\%--39.63\%.
With \texttt{DeepSeek-R1-0528}, saturation drops even more sharply---from 18.58\%--47.41\% (PQM) to 3.41\%--13.12\% (GQM).
This indicates that pointwise PQM prompting tends to over-assign the maximum score, whereas GQM's joint comparison makes differences among candidates more salient and alleviates saturation.

\paragraph{PQM-based rewards saturate early during GRPO.}

Figure~\ref{fig:rm_reward_curve} further shows that PQM-based reward modeling can lead to an early reward plateau during policy optimization.
PQM-GenRM rewards increase rapidly and approach a near-ceiling regime early in training, suggesting that the policy quickly learns to trigger near-maximal scores from the PQM reward model (i.e., reward hacking and saturation), after which the reward provides limited additional shaping.

In contrast, GRRM rewards increase more gradually and continue to fluctuate in a non-saturated regime.
This pattern is consistent with GRRM providing a more discriminative and robust signal that remains informative throughout optimization, aligning with GRPO's reliance on within-group comparisons.

\section{Quantitative Analysis of Reasoning Traces}
\label{app:reasoning_trace_analysis}

The qualitative examples in Appendix~\ref{app:mt_reasoning_cases} are complemented by a direct, issue-level evaluation of the policy reasoning traces. We use all 396 examples in Seed-X-Challenge and ask \texttt{DeepSeek-v4-Pro} to atomize the human-written expert comments into a frozen set of 1,414 translation-specific issues. The same issue set is used to evaluate the SFT checkpoint and the GRRM-optimized checkpoint independently.

\paragraph{Evaluation protocol.}
For each system, the evaluator receives only the reasoning trace; the corresponding final translation is removed. It labels every frozen issue as \textit{Correct} (the trace substantively analyzes the issue), \textit{Incorrect} (it addresses the issue but reaches an incorrect conclusion), or \textit{Missing} (the issue is not analyzed). The prompt explicitly instructs the evaluator not to reward trace length, presentation, or generic discussion. Thus, the comparison measures analysis of the same expert-identified difficulties rather than verbosity or a preference for a particular response format.

\begin{table*}[t]
    \centering
    \small
    \setlength{\tabcolsep}{2.5pt}
    \begin{tabular}{lccccccccccccc}
        \toprule
        \multirow{2}{*}{\textbf{Model}} & \multicolumn{4}{c}{\textbf{Zh$\to$En (667 issues)}} & \multicolumn{4}{c}{\textbf{En$\to$Zh (747 issues)}} & \multicolumn{4}{c}{\textbf{Overall (1,414 issues)}} \\
        \cmidrule(lr){2-5} \cmidrule(lr){6-9} \cmidrule(lr){10-13}
        & C (\%) & I (\%) & M (\%) & $\Delta$C & C (\%) & I (\%) & M (\%) & $\Delta$C & C (\%) & I (\%) & M (\%) & $\Delta$C \\
        \midrule
        Qwen2.5-7B-SFT & 61.47 & 30.88 & 7.65 & -- & 33.20 & 56.49 & 10.31 & -- & 46.53 & 44.41 & 9.05 & -- \\
        \quad + GRPO (ours) & \textbf{68.82} & 24.89 & 6.30 & \textbf{+7.35} & \textbf{51.94} & 38.15 & 9.91 & \textbf{+18.74} & \textbf{59.90} & 31.90 & 8.20 & \textbf{+13.37} \\
        \bottomrule
    \end{tabular}
    \caption{Issue-level analysis of reasoning traces on the 396 Seed-X-Challenge examples. Following the main results table, models are rows and translation directions are column groups. C/I/M denote Correct, Incorrect, and Missing percentages; $\Delta$C is the gain in Correct analysis in percentage points relative to the SFT checkpoint.}
    \label{tab:reasoning_trace_issues}
\end{table*}

As shown in Table~\ref{tab:reasoning_trace_issues}, GRRM optimization raises the overall rate of correct issue analysis by 13.37 points, with improvements in both translation directions. The transition analysis attributes most of the improvement to correcting analyses that were previously incorrect (243 issues), rather than to covering previously missing issues (44 issues). This interpretation is also consistent with the trace lengths: on En$\to$Zh, where the accuracy gain is larger, the optimized traces are substantially shorter. These results provide quantitative evidence for improved accuracy in reasoning about expert-identified translation issues.

\section{Auxiliary Score Consistency Reward}
\label{app:rm_aux_loss}

To encourage the model to generate calibrated numerical scores that reflect the magnitude of quality differences, we augment the standard ranking accuracy reward with a score consistency objective.
The GRRM is instructed to output a structured response concluding with two specific components: an explicit ranking string (e.g., $A > B = C$) and a dictionary of scalar scores (e.g., $\{A: 6, B: 5, C: 5\}$).

\paragraph{Internal Consistency Constraint.}
Before evaluating accuracy, we enforce a strict self-consistency check to ground the numerical outputs.
Let $\hat{\tau}_{\text{text}}$ be the explicit ranking string generated by the model, and $\hat{\tau}_{\text{score}}$ be the ranking induced by sorting the generated scalar scores $\hat{\mathbf{s}}$.
We define a binary consistency gate $C_{\text{gate}}$:
\begin{equation}
    C_{\text{gate}} = \mathbb{I}(\hat{\tau}_{\text{text}} \equiv \hat{\tau}_{\text{score}})
\end{equation}
This gate is applied in the reward calculation to penalize inconsistency.

\paragraph{Margin-Aware Score Reward.}
For samples passing the consistency gate, we calculate the Score Consistency Reward ($R_{\text{score}}$). Unlike simple regression losses (e.g., MSE), which can be sensitive to the absolute scale of ground-truth labels, our metric focuses on preserving the \textit{relative margins} between candidates.
Let $q_i, q_j$ be the ground-truth scores and $\hat{s}_i, \hat{s}_j$ be the predicted scores for candidates $i$ and $j$. We compute the absolute margin error $\delta_{ij}$ for every pair:
\begin{equation}
    \delta_{ij} = \left| (\hat{s}_i - \hat{s}_j) - (q_i - q_j) \right|
\end{equation}
We then map this error to a reward value using a discrete kernel function $K(\cdot)$, designed to penalize deviations while allowing for minor integer fluctuations. Based on our empirical tuning, we define:
\begin{equation}
    K(\delta) = 
    \begin{cases} 
    1.0 & \text{if } \delta = 0 \\
    0.6 & \text{if } \delta = 1 \\
    0.2 & \text{if } \delta = 2 \\
    0   & \text{otherwise}
    \end{cases}
\end{equation}
The final score consistency reward is the average kernel value across all pairs:
\begin{equation}
    R_{\text{score}} = \frac{1}{\binom{G}{2}} \sum_{1 \le i < j \le G} K(\delta_{ij})
\end{equation}

\paragraph{Total Reward.}
The final reward used for RLVR optimization is the sum of the ranking accuracy (Eq.~\ref{eq:ranking_acc}) and the score consistency reward, gated by the internal consistency check:
\begin{equation}
    R_{\text{total}} = C_{\text{gate}} \cdot (R_{\text{acc}} + R_{\text{score}})
\end{equation}
This formulation ensures that the model is optimized to produce reasoning that results in both correct ordering and precise, margin-aware numerical estimations.

\paragraph{PQM-GenRM Configuration.}
For the \textbf{PQM-GenRM} baseline, which evaluates candidates independently, we adapt the reward to target absolute accuracy. Unlike the margin-based approach in GRRM, we compute the absolute error $\delta = |\hat{s} - q|$ between the predicted score $\hat{s}$ and the ground truth $q$. The final reward utilizes a similar discrete kernel function $K(\delta)$ defined above to penalize deviations, ensuring a fair comparison between the optimization objectives.

\section{Stabilizing Group Relative Policy Optimization}
\label{app:stable_grpo}

Standard GRPO often exhibits training instability, particularly in the mid-training phase, where the model can become overly sensitive to negative rewards. This frequently leads to sharp distribution shifts and even collapse. To mitigate this, we propose \textbf{Stable GRPO}, which incorporates an auxiliary objective to anchor the policy updates to high-quality trajectories.

Recall that GRPO optimizes the policy by maximizing a clipped surrogate objective based on group relative advantages\footnote{We omit the KL regularization term for brevity.}:
{\small
\begin{equation}
\label{equ:grpo}
\mathcal{J}_\text{GRPO}(\theta) =
\mathbb{E}_{ x, \{y_i\} }
\left[ \frac{1}{G} \sum_{i=1}^{G} \frac{1}{|y_i|} \sum_{t=1}^{|y_i|}
\mathcal{L}^{\text{CLIP}}_{i,t}(\theta)
\right],
\end{equation}
}
where the clipping objective is:
{\small
\begin{equation}
\mathcal{L}^{\text{CLIP}}_{i,t}(\theta) =
\min \Big( w_{i,t}(\theta) A_{i},\,
\text{clip}\big( w_{i,t}(\theta), 1-\varepsilon, 1+\varepsilon \big) A_{i} \Big),
\end{equation}
}
and the importance ratio is:
\begin{equation}
w_{i,t}(\theta)=\frac{ \pi_{\theta} (y_{i,t} \mid x, y_{i,<t}) }
{ \pi_{\theta_\text{old}} (y_{i,t} \mid x, y_{i,<t}) }.
\end{equation}

Building upon the standard GRPO objective, we introduce an advantage-weighted SFT term:
{\small
\begin{equation}
\label{equ:stable_grpo}
\begin{aligned}
    \mathcal{J}_{\text{StableGRPO}}(\theta) = \mathbb{E}_{ x, \{y_i\} }\\
    \Bigg[ \frac{1}{G} \sum_{i=1}^{G} \frac{1}{|y_i|} \sum_{t=1}^{|y_i|} \Big( & \mathcal{L}^{\text{CLIP}}_{i,t}(\theta) + \gamma \mathcal{L}^{\text{SFT}}_{i,t}(\theta) \Big) \Bigg]
\end{aligned}
\end{equation}
}
Here, $\mathcal{L}^{\text{SFT}}_{i,t}(\theta)$ applies a standard log-likelihood maximization, strictly gated by positive advantages:
\begin{equation}
    \mathcal{L}^{\text{SFT}}_{i,t}(\theta) = \max(0, A_{i}) \ln \pi_{\theta}(y_{i,t} | x, y_{i,<t})
\end{equation}
where $\gamma$ controls the strength of this regularization.

The key design is that $\mathcal{L}^{\text{SFT}}$ is applied only to samples with positive advantages ($A_i>0$), complementing $\mathcal{L}^{\text{CLIP}}$ by strengthening updates on positive samples in the non-clipped regime and serving as an unclipped fallback signal when clipping suppresses the surrogate gradient.

We apply Stable GRPO for machine translation optimization only. For reasoning-based translation models, we set $\gamma=0.1$ (increasing to $\gamma=0.2$ for the Cross-Lingual Augmentation variant). For non-reasoning translation models, we set $\gamma=1.0$.

\section{Implementation Details}
\label{app:imp_details}

In this section, we detail the training configurations for both the Reward Model (GRRM) and the Machine Translation Policy Optimization.

\subsection{Reward Model Implementation}
\label{app:imp_rm_training}

\paragraph{Training Data Construction.}
We construct the training dataset using the Chinese-English subset of TowerBlocks (MT), by sampling group-level candidates using a seed translation model.
Specifically, we employ the \texttt{Qwen2.5-7B} model fine-tuned on the TowerBlocks (MT) dataset (approximately 150k samples). This seed model is also used as SFT baseline (w/o Reasoning) in Section~\ref{sec:mt_opt} and Appendix~\ref{app:mt_ext_ablation}.
We apply standard sampling with a temperature of $T=1.0$. For each source sentence, we sample a group of $N$ candidates, where $N \in \{2, 3, 4\}$ with a frequency ratio of 1:1:3.
To enhance candidate diversity and ensure the presence of high-quality samples within the generated groups, we inject the ground-truth reference into the candidate pool with a probability of $0.5$.
These constructed groups are subsequently annotated by \texttt{Gemini-2.5-Pro}.

\paragraph{Evaluation Data Processing.}
For the human-annotated test sets (Newstest2020 and GeneralMT2022), we calculate the final quality score for each translation candidate by averaging the ratings across raters.
From the available system outputs for each source, we retain a subset of 2 to 4 candidates, following the same distribution ratio as the training data.
However, we observe that randomly selected candidates often possess similar human quality scores (clustering near the ceiling), which introduces ambiguity into the ranking evaluation.
Therefore, to ensure a rigorous assessment of the model's ability to distinguish quality variances, we enforce a filtering constraint: the constructed group must include the candidates with the minimum and maximum human scores for that entry.

\paragraph{Training Hyperparameters.}
We employ 16 Nvidia A100 (80GB) GPUs for all training experiments detailed in this paper, including both reward modeling and the subsequent machine translation policy optimization.

In the SFT stage, we optimize the model for 3 epochs with a global batch size of 64. We employ a cosine learning rate scheduler with a peak learning rate of $6 \times 10^{-6}$ and a warmup ratio of 0.1.

In the RLVR stage, we adopt Group Sequence Policy Optimization~\citep[GSPO]{zheng2025group}, an enhanced variant of GRPO that uses sequence-level importance ratios to improve training stability and efficiency.

We set the learning rate to $1 \times 10^{-5}$ with a cosine scheduler, decaying to a minimum ratio of 0.2.
The training is conducted with a total batch size of 512 and a PPO mini-batch size of 128.
For each prompt, we generate $G=8$ rollouts with a maximum generation length of 4096 tokens. The model is trained for a single epoch.
Following recent practices in reasoning model training, we disable the KL divergence penalty to encourage broader exploration of reasoning paths.
Additionally, following Dr. GRPO~\cite{liu2025understanding}, we remove the standard deviation normalization terms from the advantage calculation in GRPO.

\subsection{MT Optimization Implementation}
\label{app:imp_mt_training}

\paragraph{SFT Hyperparameters.}
For the policy model cold-start, we perform Supervised Fine-Tuning for 1 epoch with a global batch size of 64. We utilize a cosine learning rate scheduler, setting the peak learning rate to $1 \times 10^{-5}$ with a warmup ratio of 0.1.
For the reasoning baseline, SFT uses the Zh-En data described in Section~\ref{sec:grrm_eval}. For the non-reasoning baseline used in the ablation experiments, we instead use the full multilingual TowerBlocks (MT) dataset to avoid the severe Zh-En overfitting observed under restricted non-reasoning SFT, where the model failed to generalize to other languages during RL.

\paragraph{GRPO Hyperparameters.}
We utilize GSPO and apply the same stabilization enhancement in Appendix~\ref{app:stable_grpo} on top of GSPO for MT policy optimization. The configuration largely mirrors the GRRM RLVR stage (Appendix~\ref{app:imp_rm_training}), with specific adjustments to suit the translation task.
We switch to a constant learning rate scheduler to maintain stable updates throughout the training process.
The number of rollouts per prompt is set to $G=4$, aligning with the group size capacity of our reward model.

To enhance optimization stability and reduce the gradient norm, we scale the rewards to the range $[0, 0.1]$.
Furthermore, following DAPO~\cite{yu2025dapo}, we employ a soft length penalty with the overlong buffer set to 2048 tokens.

To ensure fair comparison across different data settings, we adjust the number of training epochs to keep the total number of optimization steps consistent.
Specifically, we train for 2 epochs on the standard multilingual TowerBlocks (MT) dataset (150k samples).
For the CLA variant, we train for 1 epoch, while for the Chinese-English only subset, we extend training to 8 epochs.

\begin{figure*}[t]
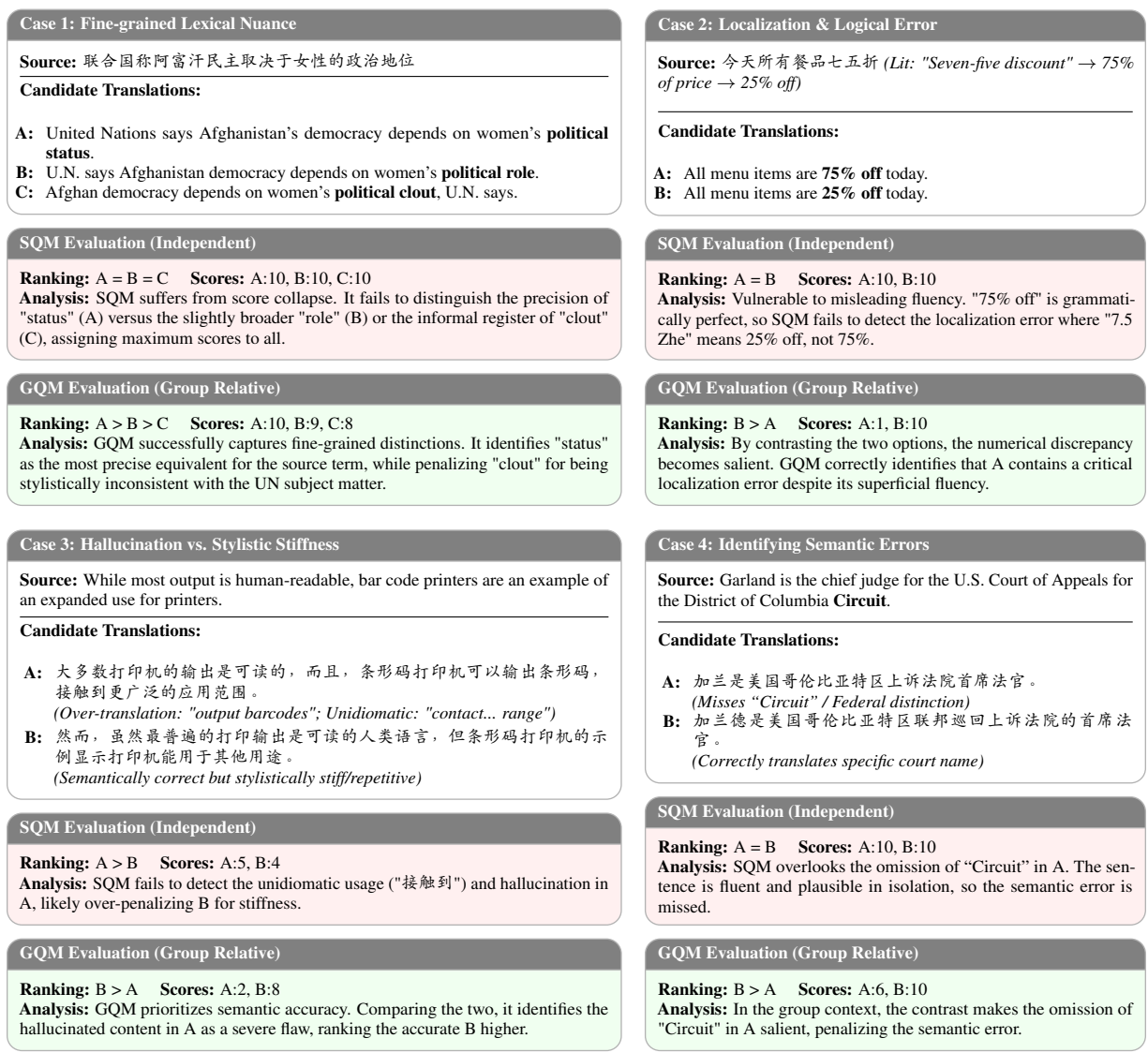

    \centering
    \scriptsize
    \definecolor{bg_input}{RGB}{245, 245, 245}
    \definecolor{bg_pqm}{RGB}{255, 240, 240} 
    \definecolor{bg_gqm}{RGB}{240, 255, 240} 
    
    \tcbset{
        colback=white,
        boxrule=0.5pt,
        colframe=gray!60,
        arc=2mm,
        left=2pt, right=2pt, top=2pt, bottom=2pt,
        fonttitle=\bfseries\small,
        title style={fill=gray!20, coltext=black}
    }

    
    \begin{minipage}[b]{0.54\textwidth}
        \begin{tcolorbox}[title={\scriptsize Case 1: Fine-grained Lexical Nuance}]
            \textbf{Source:} \begin{CJK*}{UTF8}{gkai}联合国称阿富汗民主取决于女性的政治地位\end{CJK*}
            \vspace{0.1cm}
            \hrule
            \vspace{0.1cm}
            \textbf{Candidate Translations:}
            \begin{itemize}[leftmargin=1.5em, itemsep=0pt, parsep=0pt]
                \item[\textbf{A:}] United Nations says Afghanistan's democracy depends on women's \textbf{political status}.
                \item[\textbf{B:}] U.N. says Afghanistan democracy depends on women's \textbf{political role}.
                \item[\textbf{C:}] Afghan democracy depends on women's \textbf{political clout}, U.N. says.
            \end{itemize}
        \end{tcolorbox}
        \vspace{-0.1cm}
        \begin{tcolorbox}[colback=bg_pqm, title={\scriptsize PQM Evaluation (Independent)}]
            \textbf{Ranking:} A = B = C \quad \textbf{Scores:} A:10, B:10, C:10 \\
            \textbf{Analysis:} PQM suffers from score collapse. It fails to distinguish the precision of "status" (A) versus the slightly broader "role" (B) or the informal register of "clout" (C), assigning maximum scores to all.
        \end{tcolorbox}
        \vspace{-0.1cm}
        \begin{tcolorbox}[colback=bg_gqm, title={\scriptsize GQM Evaluation (Group Relative)}]
            \textbf{Ranking:} A > B > C \quad \textbf{Scores:} A:10, B:9, C:8 \\
            \textbf{Analysis:} GQM successfully captures fine-grained distinctions. It identifies "status" as the most precise equivalent for the source term, while penalizing "clout" for being stylistically inconsistent with the UN subject matter.
        \end{tcolorbox}
    \end{minipage}
    \hfill
    \begin{minipage}[b]{0.44\textwidth}
        \begin{tcolorbox}[title={\scriptsize Case 2: Localization \& Logical Error}]
            \textbf{Source:} \begin{CJK*}{UTF8}{gkai}今天所有餐品七五折\end{CJK*} 
            \textit{(Lit: "Seven-five discount" $\to$ 75\% of price $\to$ 25\% off)}
            \vspace{0.275cm}
            \hrule
            \vspace{0.18cm}
            \textbf{Candidate Translations:}
            \begin{itemize}[leftmargin=1.5em, itemsep=0pt, parsep=0pt]
                \item[\textbf{A:}] All menu items are \textbf{75\% off} today.
                \item[\textbf{B:}] All menu items are \textbf{25\% off} today.
            \end{itemize}
        \end{tcolorbox}
        \vspace{-0.1cm}
        \begin{tcolorbox}[colback=bg_pqm, title={\scriptsize PQM Evaluation (Independent)}]
            \textbf{Ranking:} A = B \quad \textbf{Scores:} A:10, B:10 \\
            \textbf{Analysis:} Vulnerable to misleading fluency. "75\% off" is grammatically perfect, so PQM fails to detect the localization error where "7.5 Zhe" means 25\% off, not 75\%.
        \end{tcolorbox}
        \vspace{-0.1cm}
        \begin{tcolorbox}[colback=bg_gqm, title={\scriptsize GQM Evaluation (Group Relative)}]
            \textbf{Ranking:} B > A \quad \textbf{Scores:} A:1, B:10 \\
            \textbf{Analysis:} By contrasting the two options, the numerical discrepancy becomes salient. GQM correctly identifies that A contains a critical localization error despite its superficial fluency.
        \end{tcolorbox}
    \end{minipage}

    \vspace{0.3cm} 


    \begin{minipage}[b]{0.54\textwidth}
        \begin{tcolorbox}[title={\scriptsize Case 3: Hallucination vs. Stylistic Stiffness}]    
            \textbf{Source:} While most output is human-readable, bar code printers are an example of an expanded use for printers.
            \vspace{0.1cm}
            \hrule
            \vspace{0.1cm}
            \textbf{Candidate Translations:}
            \begin{itemize}[leftmargin=2em, itemsep=2pt, parsep=0pt]
                \item[\textbf{A:}] \begin{CJK*}{UTF8}{gkai}大多数打印机的输出是可读的，而且，条形码打印机可以输出条形码，接触到更广泛的应用范围。\end{CJK*} \\
                \textit{(Over-translation: "output barcodes"; Unidiomatic: "contact... range")}
                \item[\textbf{B:}] \begin{CJK*}{UTF8}{gkai}然而，虽然最普遍的打印输出是可读的人类语言，但条形码打印机的示例显示打印机能用于其他用途。\end{CJK*} \\
                \textit{(Semantically correct but stylistically stiff/repetitive)}
            \end{itemize}
        \end{tcolorbox}
        \vspace{-0.1cm}
        \begin{tcolorbox}[colback=bg_pqm, title={\scriptsize PQM Evaluation (Independent)}]
            \textbf{Ranking:} A > B \quad \textbf{Scores:} A:5, B:4 \\
            \textbf{Analysis:} PQM fails to detect the unidiomatic usage ("\begin{CJK*}{UTF8}{gkai}接触到\end{CJK*}") and hallucination in A, likely over-penalizing B for stiffness.
        \end{tcolorbox}
        \vspace{-0.1cm}
        \begin{tcolorbox}[colback=bg_gqm, title={\scriptsize GQM Evaluation (Group Relative)}]
            \textbf{Ranking:} B > A \quad \textbf{Scores:} A:2, B:8 \\
            \textbf{Analysis:} GQM prioritizes semantic accuracy. Comparing the two, it identifies the hallucinated content in A as a severe flaw, ranking the accurate B higher.
        \end{tcolorbox}
    \end{minipage}
    \hfill
    \begin{minipage}[b]{0.44\textwidth}
        \begin{tcolorbox}[title={\scriptsize Case 4: Identifying Semantic Errors}]
            \textbf{Source:} Garland is the chief judge for the U.S. Court of Appeals for the District of Columbia \textbf{Circuit}.
            \vspace{0.22cm}
            \hrule
            \vspace{0.15cm}
            \textbf{Candidate Translations:}
            \begin{itemize}[leftmargin=2em, itemsep=0pt, parsep=0pt]
                \item[\textbf{A:}] \begin{CJK*}{UTF8}{gkai}加兰是美国哥伦比亚特区上诉法院首席法官。\end{CJK*} \\
                \textit{(Misses ``Circuit'' / Federal distinction)}
                \item[\textbf{B:}] \begin{CJK*}{UTF8}{gkai}加兰德是美国哥伦比亚特区联邦巡回上诉法院的首席法官。\end{CJK*} \\
                \textit{(Correctly translates specific court name)}
            \end{itemize}
        \end{tcolorbox}
        \vspace{-0.1cm}
        \begin{tcolorbox}[colback=bg_pqm, title={\scriptsize PQM Evaluation (Independent)}]
            \textbf{Ranking:} A = B \quad \textbf{Scores:} A:10, B:10 \\
            \textbf{Analysis:} PQM overlooks the omission of ``Circuit'' in A. The sentence is fluent and plausible in isolation, so the semantic error is missed.
        \end{tcolorbox}
        \vspace{-0.1cm}
        \begin{tcolorbox}[colback=bg_gqm, title={\scriptsize GQM Evaluation (Group Relative)}]
            \textbf{Ranking:} B > A \quad \textbf{Scores:} A:6, B:10 \\
            \textbf{Analysis:} In the group context, the contrast makes the omission of "Circuit" in A salient, penalizing the semantic error.
        \end{tcolorbox}
    \end{minipage}

    \caption{Comparison of Pointwise Quality Metric (PQM) and Group Quality Metric (GQM) across four distinct scenarios. 
    \textbf{Case 1 \& 2 (Top):} Demonstrate GQM's ability to resolve fine-grained stylistic nuances and identify localization errors that PQM misses due to independent evaluation.
    \textbf{Case 3 \& 4 (Bottom):} Illustrate how GQM uses contrastive context to detect hallucinations and semantic omissions that appear fluent in isolation.
    Case 1,3,4 were conducted using \texttt{Gemini-2.5-Pro} and Case 2 was conducted using \texttt{DeepSeek-R1-0528}.}
    \label{fig:pqm_vs_gqm_cases}
\end{figure*}

\section{Independent Evaluators and Metric Divergence on Seed-X-Challenge}
\label{app:independent_evaluators}

Table~\ref{tab:mt_results} reveals a metric divergence on Seed-X-Challenge: \texttt{SeedX-PPO} obtains higher BLEURT scores than our CLA model in both directions, whereas the main LLM judge prefers our model. We therefore focus on this comparison to test whether the result depends on a particular LLM evaluator and to examine why BLEURT and LLM judgments disagree.

We conduct an additional evaluation with \texttt{GPT-5.6-Luna} and \texttt{DeepSeek-v4-Pro}. Both evaluators receive the source, human reference, and the expert comment associated with each challenge example. We evaluate the CLA variant of our policy, \texttt{SeedX-PPO}, and the controlled SFT initialization. Additionally, we use pairwise GQM to directly compare our translations against those produced by \texttt{SeedX-PPO}, reporting Win/Tie/Loss percentages from the perspective of our system.

\begin{table}[t]
\centering
\small
\setlength{\tabcolsep}{4pt}
\resizebox{\columnwidth}{!}{%
\begin{tabular}{lcccc}
\toprule
\multirow{2}{*}{\textbf{Model}} & \multicolumn{2}{c}{\textbf{GPT-5.6-Luna}} & \multicolumn{2}{c}{\textbf{DeepSeek-v4-Pro}} \\
\cmidrule(lr){2-3} \cmidrule(lr){4-5}
 & Zh$\to$En & En$\to$Zh & Zh$\to$En & En$\to$Zh \\
\midrule
SeedX-PPO & 84.78 & 86.52 & 79.78 & 79.54 \\
Qwen2.5-7B-SFT & 82.77 & 75.09 & 76.73 & 66.71 \\
\quad + GRPO w/ CLA (ours) & \textbf{89.32} & \textbf{87.18} & \textbf{85.77} & \textbf{82.10} \\
\bottomrule
\end{tabular}
}
\caption{Independent LLM-judge evaluation on Seed-X-Challenge from two additional evaluators.}
\label{tab:seedx_independent_pqm}
\end{table}

\begin{table}[t]
\centering
\small
\setlength{\tabcolsep}{5pt}
\resizebox{\columnwidth}{!}{%
\begin{tabular}{llccc}
\toprule
\textbf{Evaluator} & \textbf{Direction} & \textbf{Win} & \textbf{Tie} & \textbf{Loss} \\
\midrule
\multirow{2}{*}{GPT-5.6-Luna}
 & Zh$\to$En & \textbf{62.44} & 3.55 & 34.01 \\
 & En$\to$Zh & \textbf{48.74} & 5.03 & 46.23 \\
\midrule
\multirow{2}{*}{DeepSeek-v4-Pro}
 & Zh$\to$En & \textbf{55.84} & 4.57 & 39.59 \\
 & En$\to$Zh & \textbf{49.25} & 6.03 & 44.72 \\
\bottomrule
\end{tabular}
}
\caption{Reference-based GQM preference (\%) between Ours (+CLA) and \texttt{SeedX-PPO} on Seed-X-Challenge. Win and Loss are defined from the perspective of our system.}
\label{tab:seedx_independent_gqm}
\end{table}

As shown in Tables~\ref{tab:seedx_independent_pqm} and~\ref{tab:seedx_independent_gqm}, both evaluators reproduce the same directional pattern across the two prompting paradigms. Both place our system above \texttt{SeedX-PPO} and the SFT baseline in each direction. For \texttt{GPT-5.6-Luna}, the aggregate improvement is 2.59 points over \texttt{SeedX-PPO} and 9.33 points over SFT. Direct GQM comparison rules out independent score calibration as the sole explanation: the overall \texttt{GPT-5.6-Luna} result is 55.56\% wins, 4.29\% ties, and 40.15\% losses. The advantage is clear on Zh$\to$En, whereas En$\to$Zh is close. Thus, the overall finding is consistent across evaluators and prompting paradigms.

\paragraph{BLEURT--LLM-judge divergence.}
At the segment level, the two judges agree with each other but disagree with BLEURT on 91 examples: they prefer our translation in 71 cases and \texttt{SeedX-PPO} in 20. The latter cases show that the judges do not unconditionally favor our system.
Inspection of representative disagreements shows that the LLM judges identify critical errors involving numerical relations, terminology, semantic reversal, and hallucination in outputs from either system. This helps explain the metric divergence, while the bidirectional preferences argue against a one-sided evaluator bias.

\section{Extended Analyses}
\label{app:ext_analysis}

\begin{table*}[t]
\centering
\small
\setlength{\tabcolsep}{3.2pt}
\begin{tabular}{l c c c c c c c c}
\toprule
\multirow{2}{*}{\textbf{Model}} & \textbf{held-out} & \textbf{NT20} & \multicolumn{4}{c}{\textbf{GenMT22 (MQM)}} & \textbf{Seed-X-Challenge} & \multirow{2}{*}{\textbf{Avg.}} \\
\cmidrule(lr){2-2} \cmidrule(lr){3-3} \cmidrule(lr){4-7} \cmidrule(lr){8-8}
 & Zh$\leftrightarrow$En & Zh$\to$En & En$\to$Zh & Zh$\to$En & En$\to$De & En$\to$Ru & Zh$\leftrightarrow$En & \\
\midrule
\multicolumn{9}{l}{\textit{Additional MT evaluation baselines}} \\
MT-Ranker-XXL & 65.82 & 51.37 & 56.59 & 56.22 & 50.09 & 55.96 & 65.66 & 55.98 \\
COMET-poly & 54.13 & 46.49 & 54.78 & 49.37 & 49.64 & 54.71 & 66.60 & 53.60 \\
\midrule
\multicolumn{9}{l}{\textit{Reference rows}} \\
CometKiwi-XXL & 72.01 & 57.82 & \textbf{66.49} & 61.60 & \textbf{61.20} & \textbf{67.12} & 46.72 & 60.16 \\
BT-RM & 82.62 & \textbf{58.16} & \textbf{66.49} & 64.92 & 58.14 & 66.10 & 58.84 & 62.11 \\
\textbf{GRRM} (RLVR) & \textbf{83.19} & 57.57 & 62.12 & \textbf{65.55} & 60.75 & 66.39 & \textbf{73.93} & \textbf{64.38} \\
\bottomrule
\end{tabular}
\caption{Additional ranking accuracy (\%) with MT-Ranker and COMET-poly, compared with representative baselines from Table~\ref{tab:ranking_acc_results}.  
\textbf{Avg.} denotes the mean accuracy across external benchmarks, excluding the held-out set.}
\label{tab:ranking_acc_results_app}
\end{table*}

\subsection{Additional Contrastive Discriminative Baselines}
\label{app:contrastive_discriminative_baselines}

In addition to the main baselines discussed in Section~\ref{sec:grrm_eval}, we further compare GRRM with two recent MT evaluation models that introduce candidate-level contrastive context while still producing discriminative scores: MT-Ranker~\cite{Moosa2024MTRankerRM} and COMET-poly~\cite{zufle-etal-2025-comet}. 
This comparison is designed to disentangle the effects of two key components in our framework: (1) access to contrastive context among translation candidates, and (2) generative reasoning over the full candidate group.

Since these methods are not natively designed for our multi-candidate group ranking setting, we adapt them as follows. 
MT-Ranker is a pairwise judge; for datasets with more than two candidates per source, we aggregate pairwise outcomes into a global ordering using the ELO rating procedure. 
For Seed-X-Challenge, which is formulated as a binary ranking task, MT-Ranker is evaluated directly on pairwise comparisons. 
COMET-poly can take additional candidates as contrastive context, but it only assigns a score to the first candidate rather than producing a native list-wise ranking. We use the \texttt{COMET-poly-cand1-wmt25} model and randomly select another candidate from the same group as its contrastive context. 

Table~\ref{tab:ranking_acc_results_app} shows that contrastive discriminative evaluators outperform purely pointwise metrics in challenging two-candidate settings, but still lag behind GRRM. 
On Seed-X-Challenge, where neither ELO aggregation nor group-size mismatch is involved, MT-Ranker-XXL and COMET-poly reach 65.66\% and 66.60\% ranking accuracy, respectively---substantially above CometKiwi-XXL and BT-RM, confirming the value of candidate-level contrastive context for detecting subtle translation errors. 
However, both remain well below GRRM at 73.93\%, suggesting that contrastive context alone is not enough; generative reasoning over candidate differences provides additional gains on idioms, slang, terminology, and other difficult phenomena.

A similar pattern holds on the multi-candidate benchmarks. 
MT-Ranker-XXL and COMET-poly achieve average accuracies of 55.98\% and 53.60\%, respectively, compared with 64.38\% for GRRM. 
This drop reflects the limitation of local contrastive scoring without a global view of the full candidate set. 
ELO aggregation converts pairwise preferences into a group ranking, but cannot fully recover list-wise comparative information. 
Likewise, COMET-poly uses other candidates only as auxiliary context when scoring a single candidate, rather than jointly reasoning over the entire group.

Overall, the results suggest two conclusions: contrastive context is useful for translation reward modeling, as evidenced by the gains of MT-Ranker and COMET-poly on Seed-X-Challenge; but the full GQM formulation in GRRM is more effective, especially when GRPO requires accurate intra-group ranking over multiple policy samples. 
The consistent gap further suggests that explicit reasoning is crucial: it not only exposes candidate differences, but also helps the model make more reliable decisions on subtle and linguistically complex errors.

\begin{table*}[t]
\centering
\small
\setlength{\tabcolsep}{4.2pt}
\begin{tabular}{lccccc}
\toprule
\multirow{2}{*}{\textbf{Model}} & \multicolumn{3}{c}{\textbf{WMT Benchmarks}} & \multicolumn{2}{c}{\textbf{Seed-X-Challenge}} \\
\cmidrule(lr){2-4} \cmidrule(lr){5-6}
& \textbf{Zh$\to$En} & \textbf{En$\to$Zh} & \textbf{En$\to$X} & \textbf{Zh$\to$En} & \textbf{En$\to$Zh} \\
\midrule
\multicolumn{6}{l}{\textit{\textbf{General LLMs}}} \\
Gemini-2.5-Pro & \textbf{1.54} & \textbf{2.11} & \textbf{2.42} & \textbf{1.99} & \textbf{1.98} \\
DeepSeek-R1-0528 & 1.70 & 2.27 & 2.77 & 2.32 & 2.12 \\
Qwen3-8B & 3.47 & 4.68 & 5.30 & 4.38 & 5.13 \\
Qwen2.5-7B-Instruct & 2.13 & 3.75 & 5.43 & 3.44 & 3.54 \\
\midrule
\multicolumn{6}{l}{\textit{\textbf{Translation-Specialized Models}}} \\
TowerInstruct-13B & 2.16 & 3.18 & 3.23 & 4.50 & 3.01 \\
SeedX-PPO & \textbf{1.63} & \textbf{1.98} & \textbf{2.95} & \textbf{2.51} & \textbf{1.80} \\
SSR-X-Zero-7B & 1.80 & 2.43 & - & 2.90 & 2.48 \\
\rowcolor{gray!10} Qwen2.5-7B-SFT & 2.05 & 4.09 & 6.02 & 3.19 & 4.58 \\
\rowcolor{gray!10} \quad + GRPO (ours) & 1.68 & 2.37 & 3.41 & \textbf{2.56} & 2.37 \\
\rowcolor{gray!10} \quad + GRPO w/ CLA (ours) & \textbf{1.67} & 2.47 & 3.36 & \textbf{2.56} & 2.26 \\
\bottomrule
\end{tabular}
\caption{\textbf{MetricX-24 results on WMT and Seed-X-Challenge benchmarks.} Lower is better ($\downarrow$). The best performance within each category is highlighted in \textbf{bold}.}
\label{tab:metricx_results}
\end{table*}

\subsection{Additional Results with MetricX-24}
\label{app:metricx24_results}

To complement the main results reported with BLEURT and LLM-as-a-Judge, we further evaluate all systems using \textbf{MetricX-24} in reference-based mode.
Specifically, we use the \texttt{MetricX-24-Hybrid-XXL} evaluator in the \texttt{bfloat16} variant, producing regression-style quality scores clipped to the range \([0,25]\), where lower values indicate better translation quality. 

As shown in Table~\ref{tab:metricx_results}, the overall conclusions remain consistent with the main results. 
Among general LLMs, \texttt{Gemini-2.5-Pro} achieves the strongest MetricX-24 performance. 
Among translation-specialized models, our GRPO-based variants substantially improve over the SFT baseline across all evaluation sets, and \texttt{+GRPO w/ CLA} obtains the best or tied-best results on most settings within our model family. 
These results provide an additional validation of the gains brought by GRRM-based policy optimization.

\subsection{Full Multilingual Benchmarks}
\label{app:mt_full_results}
\begin{table*}[t]
\centering
\scriptsize 
\setlength{\tabcolsep}{1.8pt} 
\begin{tabular}{lcccccccccccccc}
\toprule
\multirow{3}{*}{\textbf{Model}} & \multicolumn{14}{c}{\textbf{WMT24++ En$\to$X Detailed Results}} \\
\cmidrule(lr){2-15}
 & \multicolumn{2}{c}{En$\to$De} & \multicolumn{2}{c}{En$\to$Es} & \multicolumn{2}{c}{En$\to$Fr} & \multicolumn{2}{c}{En$\to$It} & \multicolumn{2}{c}{En$\to$Nl} & \multicolumn{2}{c}{En$\to$Pt} & \multicolumn{2}{c}{En$\to$Ru} \\
\cmidrule(lr){2-3} \cmidrule(lr){4-5} \cmidrule(lr){6-7} \cmidrule(lr){8-9} \cmidrule(lr){10-11} \cmidrule(lr){12-13} \cmidrule(lr){14-15}
 & \tiny{BLEURT} & \tiny{R1-judge} & \tiny{BLEURT} & \tiny{R1-judge} & \tiny{BLEURT} & \tiny{R1-judge} & \tiny{BLEURT} & \tiny{R1-judge} & \tiny{BLEURT} & \tiny{R1-judge} & \tiny{BLEURT} & \tiny{R1-judge} & \tiny{BLEURT} & \tiny{R1-judge} \\
\midrule
\multicolumn{15}{l}{\textit{\textbf{General LLMs}}} \\
Gemini-2.5-Pro & \textbf{71.75} & \textbf{89.79} & \textbf{71.82} & \textbf{91.10} & \textbf{64.63} & \textbf{91.03} & \textbf{71.33} & \textbf{91.14} & \textbf{72.21} & \textbf{89.54} & \textbf{63.41} & \textbf{90.00} & \textbf{66.91} & \textbf{89.88} \\
DeepSeek-R1-0528 & 69.61 & 88.18 & 71.17 & 90.68 & 63.53 & 89.44 & 70.62 & 89.57 & 71.26 & 87.10 & 62.91 & 87.61 & 64.90 & 86.76 \\
Qwen3-8B & 63.90 & 77.13 & 64.58 & 82.71 & 40.45 & 76.21 & 62.53 & 78.98 & 62.89 & 71.37 & 56.80 & 80.58 & 49.62 & 74.62 \\
Qwen2.5-7B-Instruct & 59.43 & 68.94 & 62.84 & 78.61 & 54.47 & 76.80 & 59.84 & 71.83 & 61.44 & 64.21 & 56.64 & 76.82 & 56.41 & 70.39 \\
\midrule
\multicolumn{15}{l}{\textit{\textbf{Translation-Specialized Models}}} \\
TowerInstruct-13B & 69.11 & 82.34 & 68.68 & 84.26 & 61.51 & 83.33 & 70.35 & 84.40 & 70.49 & 82.09 & 63.37 & 83.30 & 62.73 & 79.02 \\
SeedX-PPO & \textbf{71.15} & \textbf{86.03} & \textbf{70.63} & \textbf{87.51} & \textbf{62.76} & \textbf{86.31} & \textbf{71.19} & \textbf{86.69} & \textbf{71.69} & \textbf{84.20} & \textbf{65.25} & 85.95 & \textbf{65.76} & \textbf{85.62} \\
\rowcolor{gray!10} Qwen2.5-7B-SFT & 59.06 & 65.68 & 61.91 & 74.25 & 51.33 & 71.41 & 60.12 & 66.59 & 58.10 & 60.05 & 55.77 & 72.98 & 53.72 & 64.44 \\
\rowcolor{gray!10} \quad + GRPO (ours) & 68.12 & 82.63 & 66.59 & 86.91 & 59.78 & 85.27 & 67.43 & 84.07 & 67.04 & 79.30 & 61.24 & 85.65 & 62.32 & 83.22 \\
\rowcolor{gray!10} \quad + GRPO w/ CLA (ours) & 67.92 & 82.60 & 66.44 & 86.76 & 60.18 & 85.84 & 66.98 & 83.67 & 66.50 & 78.62 & 61.12 & \textbf{86.43} & 62.39 & 82.07 \\
\bottomrule
\end{tabular}
\caption{\textbf{Detailed breakdown of WMT24++ En$\to$X results.} We report BLEURT-20 and LLM-as-a-Judge scores evaluated by \texttt{DeepSeek-R1-0528}.}
\label{tab:wmt24_detailed}
\end{table*}

\begin{table*}[t]
\centering
\small 
\setlength{\tabcolsep}{4pt}
\begin{tabular}{lcccccccc}
\toprule
\multirow{3}{*}{\textbf{Model}} & \multicolumn{8}{c}{\textbf{WMT23 X$\to$En Detailed Results}} \\
\cmidrule(lr){2-9}
 & \multicolumn{2}{c}{De$\to$En} & \multicolumn{2}{c}{Ja$\to$En} & \multicolumn{2}{c}{Ru$\to$En} & \multicolumn{2}{c}{Average} \\
\cmidrule(lr){2-3} \cmidrule(lr){4-5} \cmidrule(lr){6-7} \cmidrule(lr){8-9}
 & \scriptsize{BLEURT} & \scriptsize{oss-judge} & \scriptsize{BLEURT} & \scriptsize{oss-judge} & \scriptsize{BLEURT} & \scriptsize{oss-judge} & \scriptsize{BLEURT} & \scriptsize{oss-judge} \\
\midrule
\multicolumn{9}{l}{\textit{\textbf{General LLMs}}} \\
Qwen3-8B & 69.12 & \textbf{88.65} & 60.87 & \textbf{80.11} & 65.42 & \textbf{84.66} & 65.14 & \textbf{84.47} \\
Qwen2.5-7B-Instruct & \textbf{71.75} & 86.21 & \textbf{66.23} & 79.03 & \textbf{71.04} & 83.80 & \textbf{69.67} & 83.01 \\
\midrule
\multicolumn{9}{l}{\textit{\textbf{Translation-Specialized Models}}} \\
TowerInstruct-13B & 74.48 & 90.04 & 66.74 & 75.66 & 72.98 & 86.46 & 71.40 & 84.06 \\
SeedX-PPO & \textbf{75.43} & \textbf{92.05} & \textbf{70.19} & 84.99 & \textbf{74.40} & 89.82 & \textbf{73.34} & \textbf{88.95} \\
\rowcolor{gray!10} Qwen2.5-7B-SFT & 71.51 & 85.54 & 66.38 & 78.08 & 71.15 & 84.15 & 69.68 & 82.59 \\
\rowcolor{gray!10} \quad + GRPO (ours) & 70.25 & 90.69 & 67.72 & \textbf{85.50} & 71.93 & \textbf{90.14} & 69.97 & 88.78 \\
\rowcolor{gray!10} \quad + GRPO w/ CLA (ours) & 70.83 & 90.47 & 67.73 & 84.98 & 71.83 & 89.25 & 70.13 & 88.23 \\
\bottomrule
\end{tabular}
\caption{\textbf{Detailed breakdown of WMT23 X$\to$En results.} We report BLEURT-20 and LLM-as-a-Judge scores evaluated by \texttt{gpt-oss-120b}.}
\label{tab:wmt23_detailed}
\end{table*}

We provide the detailed breakdown of the aggregated performance reported in the main results. Table~\ref{tab:wmt24_detailed} details the En$\to$X performance across seven European and Slavic languages from WMT24++, while Table~\ref{tab:wmt23_detailed} presents the WMT23 X$\to$En results for German, Japanese, and Russian.

The results demonstrate that our GRPO-based optimization yields consistent improvements over the SFT baseline across diverse language families. For WMT23 X$\to$En benchmark, while \texttt{SeedX-PPO} generally leads in BLEURT, our method achieves superior or competitive LLM judge scores, particularly in directions like Ja$\to$En and Ru$\to$En.

\subsection{Extended Ablation Studies}
\label{app:mt_ext_ablation}

\begin{table*}[t]
\centering
\small 
\setlength{\tabcolsep}{3.2pt}
\begin{tabular}{lcccccccccc}
\toprule
\multirow{3}{*}{\textbf{Supplementary Configuration}} & \multicolumn{6}{c}{\textbf{WMT Benchmarks}} & \multicolumn{4}{c}{\textbf{Seed-X-Challenge}} \\
\cmidrule(lr){2-7} \cmidrule(lr){8-11}
 & \multicolumn{2}{c}{Zh$\to$En} & \multicolumn{2}{c}{En$\to$Zh} & \multicolumn{2}{c}{En$\to$X} & \multicolumn{2}{c}{Zh$\to$En} & \multicolumn{2}{c}{En$\to$Zh} \\
\cmidrule(lr){2-3} \cmidrule(lr){4-5} \cmidrule(lr){6-7} \cmidrule(lr){8-9} \cmidrule(lr){10-11}
 & \scriptsize{BLEURT} &\scriptsize{oss-judge} & \scriptsize{BLEURT} &\scriptsize{oss-judge} & \scriptsize{BLEURT} &\scriptsize{oss-judge} & \scriptsize{BLEURT} &\scriptsize{oss-judge} & \scriptsize{BLEURT} &\scriptsize{oss-judge} \\
\midrule
\multicolumn{11}{l}{\textit{\textbf{Comparison of Reward Models for Non-Reasoning Baselines}}} \\
SFT baseline (w/o Reasoning) & 67.61 & 85.28 & 63.50 & 79.42 & 63.24 & 75.33 & 65.52 & 76.83 & 64.04 & 73.34 \\
\quad + GRPO w/ BLEURT & 68.36 & 86.46 & 65.28 & 81.19 & 65.12 & 77.49 & 67.21 & 79.17 & 65.25 & 75.99 \\
\quad + GRPO w/ BT-RM & 68.26 & 88.35 & 64.67 & 85.44 & 63.33 & 78.21 & 69.41 & 82.88 & 66.00 & 78.85 \\
\quad + GRPO w/ PQM-GenRM & 68.02 & 87.55 & 38.88 & 36.08 & 27.14 & 7.13 & 68.32 & 82.10 & 42.49 & 39.85 \\
\rowcolor{gray!10} \quad + GRPO w/ GRRM & 67.38 & 88.42 & 64.15 & 85.19 & 63.40 & 78.88 & 68.57 & 85.38 & 65.92 & 79.35 \\
\midrule
\multicolumn{11}{l}{\textit{\textbf{GRRM with Instruct-Tuned Base Model}}} \\
Qwen2.5-7B-Instruct & 67.31 & 85.84 & 59.92 & 79.42 & 58.72 & 70.84 & 66.59 & 79.95 & 62.75 & 73.93 \\
\rowcolor{gray!10} \quad + GRPO w/ GRRM & 67.15 & 89.40 & 64.88 & 87.07 & 63.89 & 82.44 & 68.74 & 86.99 & 67.37 & 83.45 \\
\bottomrule
\end{tabular}
\caption{\textbf{Additional ablation results.} We report performance for (1) Applying different reward models to a non reasoning model tuned from \texttt{Qwen2.5-7B}, and (2) Applying GRRM optimization on top of the instruction-tuned \texttt{Qwen2.5-7B-Instruct} model.}
\label{tab:app_ablation}
\end{table*}

In this section, we provide supplementary experiments to further validate the robustness and universality of our proposed framework. Table~\ref{tab:app_ablation} presents the detailed results.

\paragraph{Reward Model Stability on Non-Reasoning Baselines.}
We first investigate whether the instability of certain reward models (observed in Section~\ref{sec:mt_ablation}) persists when optimizing a standard, non-reasoning translation policy.
As shown in the first block of Table~\ref{tab:app_ablation}, the \textbf{PQM-GenRM} continues to exhibit severe failure modes, particularly on En$\to$X tasks (dropping to a score of 7.13), confirming that the vulnerability to reward hacking is inherent to the reward model itself rather than specific to reasoning policies.
In contrast, our \textbf{GRRM} consistently outperforms other reward signals (BLEURT, BT-RM) across most benchmarks.
However, it is worth noting that even with the best reward model (GRRM), the non-reasoning baseline (Score 79.35 on Seed-X En$\to$Zh) still significantly lags behind the reasoning-enhanced policy reported in the main text (Score 84.36). This further corroborates our conclusion that the "Reasoning $\times$ Reasoning" paradigm is essential for peak performance.

\paragraph{Generalization to Instruction-Tuned Models.}
To assess the universality of our approach, we apply our GRPO training with GRRM directly to a standard instruction-tuned model, \texttt{Qwen2.5-7B-Instruct}, without any task-specific SFT warm-up.
As shown in the second block of Table~\ref{tab:app_ablation}, our method yields substantial improvements over the strong base model.
Notably, on the unseen En$\to$X task, the performance improves by over 11 points on the \texttt{oss-judge} metric (70.84 $\to$ 82.44).
This result demonstrates that our framework is not limited to specialized translation models but can serve as a general-purpose alignment technique to enhance the multilingual capabilities of off-the-shelf LLMs.

\subsection{Case Studies: Reasoning on Challenging Translation Cases}
\label{app:mt_reasoning_cases}

\begin{figure*}[t]
    \centering
    \scriptsize
    \definecolor{bg_input}{RGB}{245, 245, 245}
    \definecolor{bg_reasoning}{RGB}{242, 248, 255} 
    \definecolor{bg_output}{RGB}{240, 255, 240}    
    
    \tcbset{
        colback=white,
        boxrule=0.5pt,
        colframe=gray!60,
        arc=2mm,
        left=3pt, right=3pt, top=3pt, bottom=3pt,
        fonttitle=\bfseries\small,
        title style={fill=gray!20, coltext=black}
    }

    
    \begin{minipage}[b]{0.8\textwidth}
        \begin{tcolorbox}[colback=bg_input, title={\scriptsize GRRM Prompt Example}]
            \textbf{User:} Given a source text in English and multiple translation candidates in Chinese. Perform a step by step analysis and comparison of the translation quality for the candidates. Finally, rank and score the candidates with integer scores on a scale from 0 to 10.
            \newline
            \newline
            Source text:
            \newline
            \codefence
            \newline
            Garland is the chief judge for the U.S. Court of Appeals for the District of Columbia Circuit.
            \newline
            \codefence
            \newline
            \newline
            Translation A:
            \newline
            \codefence\newline
            \begin{CJK*}{UTF8}{gkai}加兰是美国哥伦比亚特区上诉法院首席法官。\end{CJK*}
            \newline
            \codefence\newline
            Translation B:
            \newline
            \codefence\newline
            \begin{CJK*}{UTF8}{gkai}加兰德是美国哥伦比亚特区联邦巡回上诉法院的首席法官。\end{CJK*}
            \newline
            \codefence\newline
        \end{tcolorbox}
        \vspace{-0.15cm}
        
        \begin{tcolorbox}[colback=bg_reasoning, title={\scriptsize GRRM Output}]
            \#\#\# Analysis and Comparison
            \newline
            \newline
            \textbf{Source Text:} Garland is the chief judge for the U.S. Court of Appeals for the District of Columbia Circuit.
            \newline
            \newline
            1.  \textbf{Translation A: \begin{CJK*}{UTF8}{gkai}加兰是美国哥伦比亚特区上诉法院首席法官。\end{CJK*}}
            \newline
            \hspace*{1em} *   \textbf{Accuracy:} This translation correctly identifies Garland as the chief judge of a court in the D.C. area, but it is imprecise. The "Court of Appeals for the District of Columbia Circuit" is a specific federal appellate court in the U.S. legal system. "District of Columbia Circuit" is a key part of its name that signifies it's part of a "Circuit" of federal appeals courts, not just any court in D.C. By translating "Circuit" as "\begin{CJK*}{UTF8}{gkai}上诉法院\end{CJK*}" (Court of Appeals), the translation implies he is a chief judge of the D.C. "appellate court" generally, which misses the "Circuit" designation that distinguishes it from other appellate courts. This is a loss of specific legal precision.
            \newline
            \hspace*{1em} *   \textbf{Clarity:} The meaning is generally clear.
            \newline
            \newline
            2.  \textbf{Translation B: \begin{CJK*}{UTF8}{gkai}加兰德是美国哥伦比亚特区联邦巡回上诉法院的首席法官。\end{CJK*}}
            \newline
            \hspace*{1em} *   \textbf{Accuracy:} This translation is highly accurate. "\begin{CJK*}{UTF8}{gkai}联邦巡回上诉法院\end{CJK*}" (Federal Circuit Court of Appeals) is the standard official name for "U.S. Court of Appeals" in the context of U.S. federal courts. The structure "Circuit Court of Appeals for the District of Columbia" implies it's the D.C. court of this specific federal judicial "Circuit". This translation correctly identifies Garland as the chief judge of the full, specific court with the correct legal terminology in Chinese.
            \newline
            \hspace*{1em} *   \textbf{Clarity:} The meaning is unambiguous and fully preserves the nuances of the original text. "\begin{CJK*}{UTF8}{gkai}加兰德\end{CJK*}" is the standard pinyin.
            \newline
            \newline
            \#\#\# Conclusion
            \newline
            \newline
            Translation A is acceptable but lacks specificity due to the omission of "Circuit" in the court's name. It's understandable but not technically correct in a legal context.
            \newline
            \newline
            Translation B is excellent. It uses the precise and official name "\begin{CJK*}{UTF8}{gkai}联邦巡回上诉法院\end{CJK*}" for the "U.S. Court of Appeals", correctly translating "Circuit" as part of the full name of the court. It is both accurate and clear.
            \newline
            \newline
            \#\#\# Final Ranking:
            \newline
            \newline
            B > A
            \newline
            \newline
            \#\#\# Scores:
            \newline
            \newline
            B: 10, A: 7
            
        \end{tcolorbox}
        
    \end{minipage}

    \caption{An example of GRRM for machine translation evaluation. Given a source sentence and multiple translation candidates, the model performs comparative reasoning, then outputs a ranking and relative integer scores. The GQM prompt used for data annotation is similar, except for the additional requirements regarding the output format.}

    \label{fig:GRRM_example}
\end{figure*}

\begin{figure*}[t]
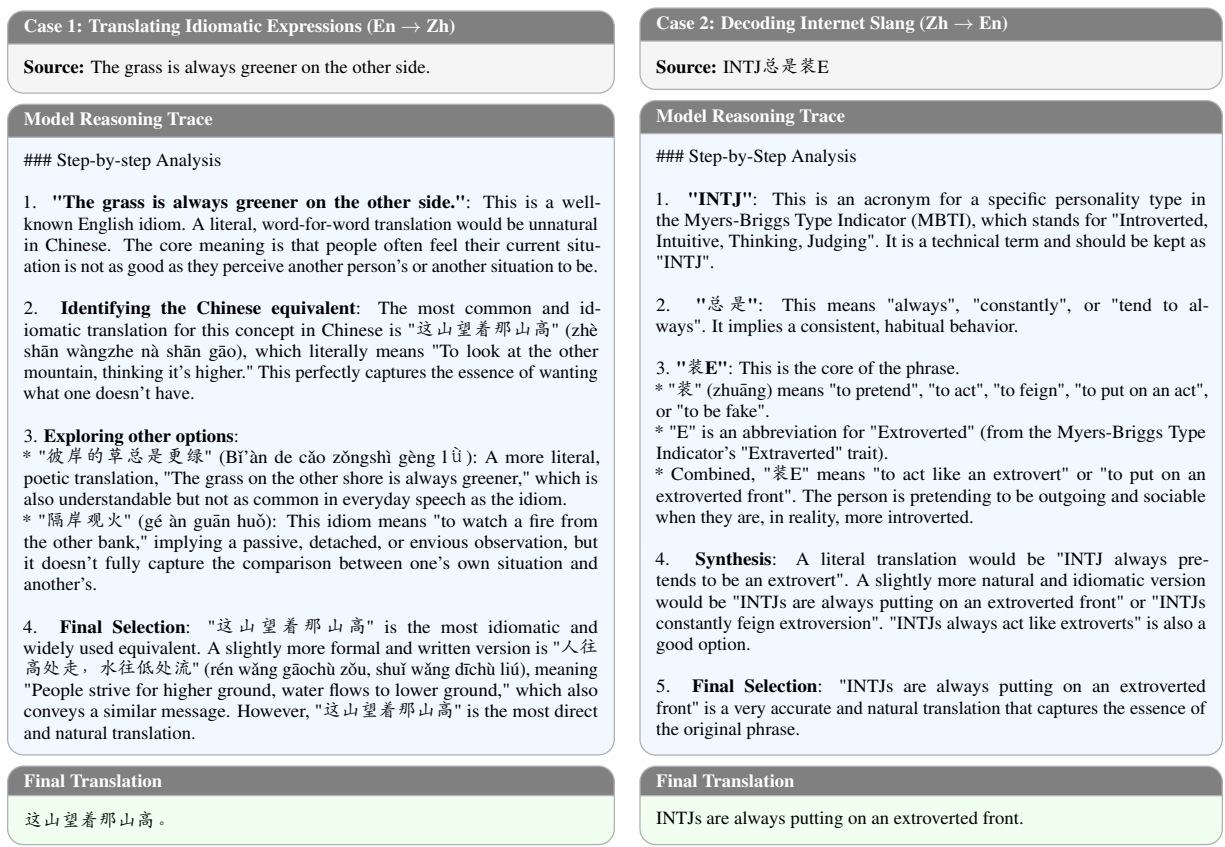

    \centering
    \scriptsize
    \definecolor{bg_input}{RGB}{245, 245, 245}
    \definecolor{bg_reasoning}{RGB}{242, 248, 255} 
    \definecolor{bg_output}{RGB}{240, 255, 240}    
    
    \tcbset{
        colback=white,
        boxrule=0.5pt,
        colframe=gray!60,
        arc=2mm,
        left=3pt, right=3pt, top=3pt, bottom=3pt,
        fonttitle=\bfseries\small,
        title style={fill=gray!20, coltext=black}
    }

    
    \begin{minipage}[b]{0.5\textwidth}
        \begin{tcolorbox}[colback=bg_input, title={\scriptsize Case 1: Translating Idiomatic Expressions (En $\to$ Zh)}]
            \textbf{Source:} The grass is always greener on the other side.
        \end{tcolorbox}
        \vspace{-0.15cm}
        
        \begin{tcolorbox}[colback=bg_reasoning, title={\scriptsize Model Reasoning Trace}]
            \#\#\# Step-by-step Analysis
            \newline
            \newline
            1.  \textbf{"The grass is always greener on the other side."}: This is a well-known English idiom. A literal, word-for-word translation would be unnatural in Chinese. The core meaning is that people often feel their current situation is not as good as they perceive another person's or another situation to be.
            \newline
            \newline
            2.  \textbf{Identifying the Chinese equivalent}: The most common and idiomatic translation for this concept in Chinese is "\begin{CJK*}{UTF8}{gkai}这山望着那山高\end{CJK*}" (zhè shān wàngzhe nà shān gāo), which literally means "To look at the other mountain, thinking it's higher." This perfectly captures the essence of wanting what one doesn't have.
            \newline
            \newline
            3.  \textbf{Exploring other options}:
                \newline
                \hspace*{1em} *   "\begin{CJK*}{UTF8}{gkai}彼岸的草总是更绿\end{CJK*}" (\begin{CJK*}{UTF8}{gbsn}Bǐ'àn de cǎo zǒngshì gèng lǜ\end{CJK*}): A more literal, poetic translation, "The grass on the other shore is always greener," which is also understandable but not as common in everyday speech as the idiom.
                \newline
                \hspace*{1em} *   "\begin{CJK*}{UTF8}{gkai}隔岸观火\end{CJK*}" (gé àn guān huǒ): This idiom means "to watch a fire from the other bank," implying a passive, detached, or envious observation, but it doesn't fully capture the comparison between one's own situation and another's.
            \newline
            \newline
            4.  \textbf{Final Selection}: "\begin{CJK*}{UTF8}{gkai}这山望着那山高\end{CJK*}" is the most idiomatic and widely used equivalent. A slightly more formal and written version is "\begin{CJK*}{UTF8}{gkai}人往高处走，水往低处流\end{CJK*}" (rén wǎng gāochù zǒu, shuǐ wǎng dīchù liú), meaning "People strive for higher ground, water flows to lower ground," which also conveys a similar message. However, "\begin{CJK*}{UTF8}{gkai}这山望着那山高\end{CJK*}" is the most direct and natural translation.
        \end{tcolorbox}
        \vspace{-0.15cm}
        
        \begin{tcolorbox}[colback=bg_output, title={\scriptsize Final Translation}]
            \begin{CJK*}{UTF8}{gkai}这山望着那山高。\end{CJK*}
        \end{tcolorbox}
    \end{minipage}
    \hfill
    \begin{minipage}[b]{0.48\textwidth}
        \begin{tcolorbox}[colback=bg_input, title={\scriptsize Case 2: Decoding Internet Slang (Zh $\to$ En)}]
            \textbf{Source:} \begin{CJK*}{UTF8}{gkai}INTJ总是装E\end{CJK*}
        \end{tcolorbox}
        \vspace{-0.15cm}
        
        \begin{tcolorbox}[colback=bg_reasoning, title={\scriptsize Model Reasoning Trace}]
            \#\#\# Step-by-Step Analysis
            \newline
            \newline
            1.  \textbf{"INTJ"}: This is an acronym for a specific personality type in the Myers-Briggs Type Indicator (MBTI), which stands for "Introverted, Intuitive, Thinking, Judging". It is a technical term and should be kept as "INTJ".
            \newline
            \newline
            2.  \textbf{"\begin{CJK*}{UTF8}{gkai}总是\end{CJK*}"}: This means "always", "constantly", or "tend to always". It implies a consistent, habitual behavior.
            \newline
            \newline
            3.  \textbf{"\begin{CJK*}{UTF8}{gkai}装E\end{CJK*}"}: This is the core of the phrase.
                \newline
                \hspace*{1em} *   "\begin{CJK*}{UTF8}{gkai}装\end{CJK*}" (zhuāng) means "to pretend", "to act", "to feign", "to put on an act", or "to be fake".
                \newline
                \hspace*{1em} *   "E" is an abbreviation for "Extroverted" (from the Myers-Briggs Type Indicator's "Extraverted" trait).
                \newline
                \hspace*{1em} *   Combined, "\begin{CJK*}{UTF8}{gkai}装E\end{CJK*}" means "to act like an extrovert" or "to put on an extroverted front". The person is pretending to be outgoing and sociable when they are, in reality, more introverted.
            \newline
            \newline
            4.  \textbf{Synthesis}: A literal translation would be "INTJ always pretends to be an extrovert". A slightly more natural and idiomatic version would be "INTJs are always putting on an extroverted front" or "INTJs constantly feign extroversion". "INTJs always act like extroverts" is also a good option.
            \newline
            \newline
            5.  \textbf{Final Selection}: "INTJs are always putting on an extroverted front" is a very accurate and natural translation that captures the essence of the original phrase.
        \end{tcolorbox}
        \vspace{-0.15cm}
        
        \begin{tcolorbox}[colback=bg_output, title={\scriptsize Final Translation}]
            INTJs are always putting on an extroverted front.
        \end{tcolorbox}
    \end{minipage}

    \caption{Qualitative analysis of translation reasoning. We showcase the model's ability to handle cultural nuances by identifying \textbf{idiomatic equivalents} (Case 1) and \textbf{decoding internet slang} (Case 2) through step-by-step analysis.}

    \label{fig:reasoning_cases}
\end{figure*}

To provide a qualitative perspective on the quantitative gains observed in the Seed-X-Challenge, Figure~\ref{fig:reasoning_cases} illustrates how our GRRM-optimized model reasons through challenging translation cases involving cultural ambiguities.

In the scenario of \textbf{Translating Idiomatic Expressions} (Case 1), the model successfully navigates the English proverb ``The grass is always greener on the other side.'' Instead of producing a rigid or literal translation (e.g., regarding the color of grass), the model identifies the underlying semantic meaning of dissatisfaction and envy. It then retrieves the culturally equivalent Chinese proverb ``\begin{CJK*}{UTF8}{gkai}这山望着那山高\end{CJK*}'' (\textit{This mountain looks higher than that one}), ensuring the translation resonates with native speakers.

Similarly, in the \textbf{Decoding Internet Slang} (Case 2), the model encounters the phrase ``\begin{CJK*}{UTF8}{gkai}INTJ总是装E\end{CJK*}'', which blends technical MBTI terminology with colloquial Chinese slang. A literal breakdown might fail to capture the nuance of ``\begin{CJK*}{UTF8}{gkai}装\end{CJK*}'' (feign/pretend) in this specific context. However, the model's reasoning trace explicitly decomposes the acronyms and analyzes the character traits, correctly deriving the idiomatic English translation: ``putting on an extroverted front.'' 

These cases confirm that our approach does not merely memorize translation pairs but actively performs semantic analysis and cultural alignment, justifying the substantial improvements seen in the LLM-judge evaluations.

\section{GRRM Application: Inference-time Reranking}
\label{app:rm_inference_reranking}

\begin{table*}[t]
\centering
\scriptsize
\setlength{\tabcolsep}{2.5pt} 
\begin{tabular}{llcccccccccc}
\toprule
\multirow{3}{*}{\textbf{Base Model}} & \multirow{3}{*}{\textbf{Inference Strategy}} & \multicolumn{2}{c}{\textbf{WMT23}} & \multicolumn{2}{c}{\textbf{WMT24++}} & \multicolumn{4}{c}{\textbf{Seed-X-Challenge}} & \multicolumn{2}{c}{\multirow{2}{*}{\textbf{Average}}} \\
\cmidrule(lr){3-4} \cmidrule(lr){5-6} \cmidrule(lr){7-10} 
 & & \multicolumn{2}{c}{Zh$\to$En} & \multicolumn{2}{c}{En$\to$Zh} & \multicolumn{2}{c}{Zh$\to$En} & \multicolumn{2}{c}{En$\to$Zh} & \multicolumn{2}{c}{} \\
\cmidrule(lr){3-3} \cmidrule(lr){4-4} \cmidrule(lr){5-5} \cmidrule(lr){6-6} \cmidrule(lr){7-7} \cmidrule(lr){8-8} \cmidrule(lr){9-9} \cmidrule(lr){10-10} \cmidrule(lr){11-12}
 & & \scriptsize{BLEURT} & \scriptsize{oss-judge} & \scriptsize{BLEURT} & \scriptsize{oss-judge} & \scriptsize{BLEURT} & \scriptsize{oss-judge} & \scriptsize{BLEURT} & \scriptsize{oss-judge} & \scriptsize{BLEURT} & \scriptsize{oss-judge} \\
\midrule

\multirow{3}{*}{\shortstack[l]{SFT Baseline\\(w/ Reasoning)}} 
 & Sampling Baseline & 66.78 & 85.04 & 59.33 & 74.20 & 66.95 & 80.17 & 61.64 & 70.02 & 63.67 & 77.36 \\
 & Best-of-$N$ & 66.76 & 85.21 & 59.79 & 75.50 & 66.88 & 79.57 & 61.94 & 71.27 & 63.84 & 77.89 \\
 & \textbf{Ranking w/ GRRM} & \textbf{67.68} & \textbf{88.06} & \textbf{61.50} & \textbf{81.47} & \textbf{68.05} & \textbf{83.22} & \textbf{63.29} & \textbf{77.15} & \textbf{65.13} & \textbf{82.47} \\
\midrule

\multirow{3}{*}{\shortstack[l]{+ GRPO w/ GRRM}} 
 & Sampling Baseline & 67.40 & 90.15 & 64.77 & 88.05 & 69.23 & 87.41 & 67.20 & 84.79 & 67.15 & 87.60 \\
 & Best-of-$N$ & \textbf{67.55} & 90.28 & 64.99 & 88.26 & \textbf{69.42} & 87.73 & 67.21 & 83.89 & 67.29 & 87.54 \\
 & \textbf{Ranking w/ GRRM} & 67.45 & \textbf{90.41} & \textbf{65.17} & \textbf{88.90} & 69.16 & \textbf{87.77} & \textbf{67.54} & \textbf{85.90} & \textbf{67.33} & \textbf{88.25} \\
\midrule

\multirow{4}{*}{\shortstack[l]{SFT Baseline\\(w/o Reasoning)}} 
 & Sampling Baseline & 66.05 & 82.63 & 61.50 & 74.63 & 64.16 & 73.75 & 63.04 & 69.29 & 63.69 & 75.07 \\
 & Best-of-$N$ & 67.11 & 84.49 & 62.42 & 77.48 & 64.88 & 73.85 & 62.91 & 71.00 & 64.33 & 76.71 \\
 & Beam Search & \textbf{68.18} & 86.28 & \textbf{63.62} & \textbf{81.34} & \textbf{66.25} & 77.68 & \textbf{64.16} & 75.45 & \textbf{65.55} & 80.19 \\
 & \textbf{Ranking w/ GRRM} & 67.54 & \textbf{86.65} & 63.18 & 81.06 & 65.75 & \textbf{78.35} & 63.99 & \textbf{75.46} & 65.12 & \textbf{80.38} \\
\midrule

\multirow{4}{*}{\shortstack[l]{+ GRPO w/ GRRM}} 
 & Sampling Baseline & 67.17 & 88.10 & 63.91 & 84.93 & 68.19 & 84.97 & 65.96 & 79.44 & 66.31 & 84.36 \\
 & Best-of-$N$ & 67.38 & 88.47 & 64.12 & 85.11 & 68.47 & 85.24 & 65.66 & 78.00 & 66.40 & 84.21 \\
 & Beam Search & \textbf{67.54} & 88.81 & 64.20 & 85.45 & 68.48 & 85.62 & 65.98 & 78.64 & 66.55 & 84.63 \\
 & \textbf{Ranking w/ GRRM} & 67.37 & \textbf{89.21} & \textbf{64.32} & \textbf{86.27} & \textbf{68.53} & \textbf{86.59} & \textbf{66.60} & \textbf{80.03} & \textbf{66.70} & \textbf{85.53} \\
\bottomrule
\end{tabular}
\caption{\textbf{Inference-time Reranking Performance.} We compare different decoding strategies across reasoning and non-reasoning models.}
\label{tab:reranking_full}
\end{table*}

In this section, we investigate the effectiveness of GRRM as a verifier to select the best translation from multiple candidates during inference.

\paragraph{Experimental Setup.}
We conduct experiments on both reasoning and non-reasoning models, evaluating their respective SFT baselines and GRPO-optimized checkpoints.
For candidate generation, we employ standard sampling with a temperature of $0.6$ to produce $N=4$ candidates per input.
We compare our GRRM reranking strategy against:
(1) \textbf{Sampling Baseline}, representing the expected performance of the policy;
and (2) \textbf{Best-of-$N$}, which selects the candidate with the highest log-probability assigned by the policy model.
Additionally, for non-reasoning models, we include \textbf{Beam Search} with a beam width of 4 as a strong decoding baseline.
Notably, we exclude Beam Search for reasoning models, as the extensive length of Chain-of-Thought sequences renders the decoding process prohibitively slow and computationally intractable.

\paragraph{Results and Analysis.}
The results presented in Table~\ref{tab:reranking_full} show that ranking with GRRM consistently outperforms the Sampling and Best-of-$N$ baselines across all datasets and model stages. Even for GRPO-optimized models, which have already been aligned via RL, inference-time reranking provides further performance gains.
For reasoning models where Beam Search is computationally intractable, \emph{scaling test-time compute} via GRRM reranking proves to be a robust inference paradigm. Rather than relying on a single generation, this verify-and-select strategy significantly boosts the SFT baseline (e.g., an average increase of over 5 points in LLM-judge scores) by effectively filtering out flawed reasoning paths.
In non-reasoning settings, while Beam Search remains competitive on BLEURT, GRRM reranking consistently achieves superior LLM-judge scores. This suggests that GRRM prioritizes semantic fidelity and better aligns with human preferences compared to likelihood-based decoding strategies.